\documentclass[10pt,twocolumn,letterpaper]{article}

\usepackage{iccv}
\usepackage{times}
\usepackage{epsfig}
\usepackage{graphicx}
\usepackage{amsmath}
\usepackage{amssymb}
\usepackage{booktabs}
\usepackage{multirow}
\usepackage{subcaption}
\usepackage{appendix}
\usepackage[pagebackref=true,breaklinks=true,letterpaper=true,colorlinks,bookmarks=false]{hyperref}

 \iccvfinalcopy % *** Uncomment this line for the final submission

\def\iccvPaperID{6268} % *** Enter the ICCV Paper ID here

\ificcvfinal\fi

\begin{document}

%%%%%%%%% TITLE
\title{Spatial Self-Distillation for Object Detection with Inaccurate Bounding Boxes}

% \author{First Author\\
% Institution1\\
% Institution1 address\\
% {\tt\small firstauthor@i1.org}
% % For a paper whose authors are all at the same institution,
% % omit the following lines up until the closing ``}''.
% % Additional authors and addresses can be added with ``\and'',
% % just like the second author.
% % To save space, use either the email address or home page, not both
% \and
% Second Author\\
% Institution2\\
% First line of institution2 address\\
% {\tt\small secondauthor@i2.org}
% }
\author{Di Wu\thanks{\ Equal contribution.} , Pengfei Chen$^{*}$, Xuehui Yu$^{*}$, Guorong Li, % \\ 
Zhenjun Han\thanks{\ Corresponding authors. (hanzhj@ucas.ac.cn)} , Jianbin Jiao \\
{\tt\small University of Chinese Academy of Sciences}
}

\maketitle
% Remove page # from the first page of camera-ready.
\ificcvfinal\thispagestyle{empty}\fi

%%%%%%%%% ABSTRACT
\begin{abstract}
    Object detection via inaccurate bounding boxes supervision has boosted a broad interest due to the expensive high-quality annotation data or the occasional inevitability of low annotation quality (\eg tiny objects).
    The previous works usually utilize multiple instance learning (MIL), which highly depends on category information, to select and refine a low-quality box. Those methods suffer from object drift, group prediction and part domination problems without exploring spatial information.
    In this paper, we heuristically propose a \textbf{Spatial Self-Distillation based Object Detector (SSD-Det)} to mine spatial information to refine the inaccurate box in a self-distillation fashion. SSD-Det utilizes a Spatial Position Self-Distillation \textbf{(SPSD)} module to exploit spatial information and an interactive structure to combine spatial information and category information, thus constructing a high-quality proposal bag. To further improve the selection procedure, a Spatial Identity Self-Distillation \textbf{(SISD)} module is introduced in SSD-Det to obtain spatial confidence to help select the best proposals.
    Experiments on MS-COCO and VOC datasets with noisy box annotation verify our method's effectiveness and achieve state-of-the-art performance. The code is available at \url{https://github.com/ucas-vg/PointTinyBenchmark/tree/SSD-Det}.
\end{abstract}

%%%%%%%%% BODY TEXT 
\section{Introduction}

\begin{figure}[t]
  \centering
  \subcaptionbox{Labelling strategies lead to inaccurate annotations (red box).\label{fig: noisy source}}{
  \centering
  \includegraphics[width=\linewidth]{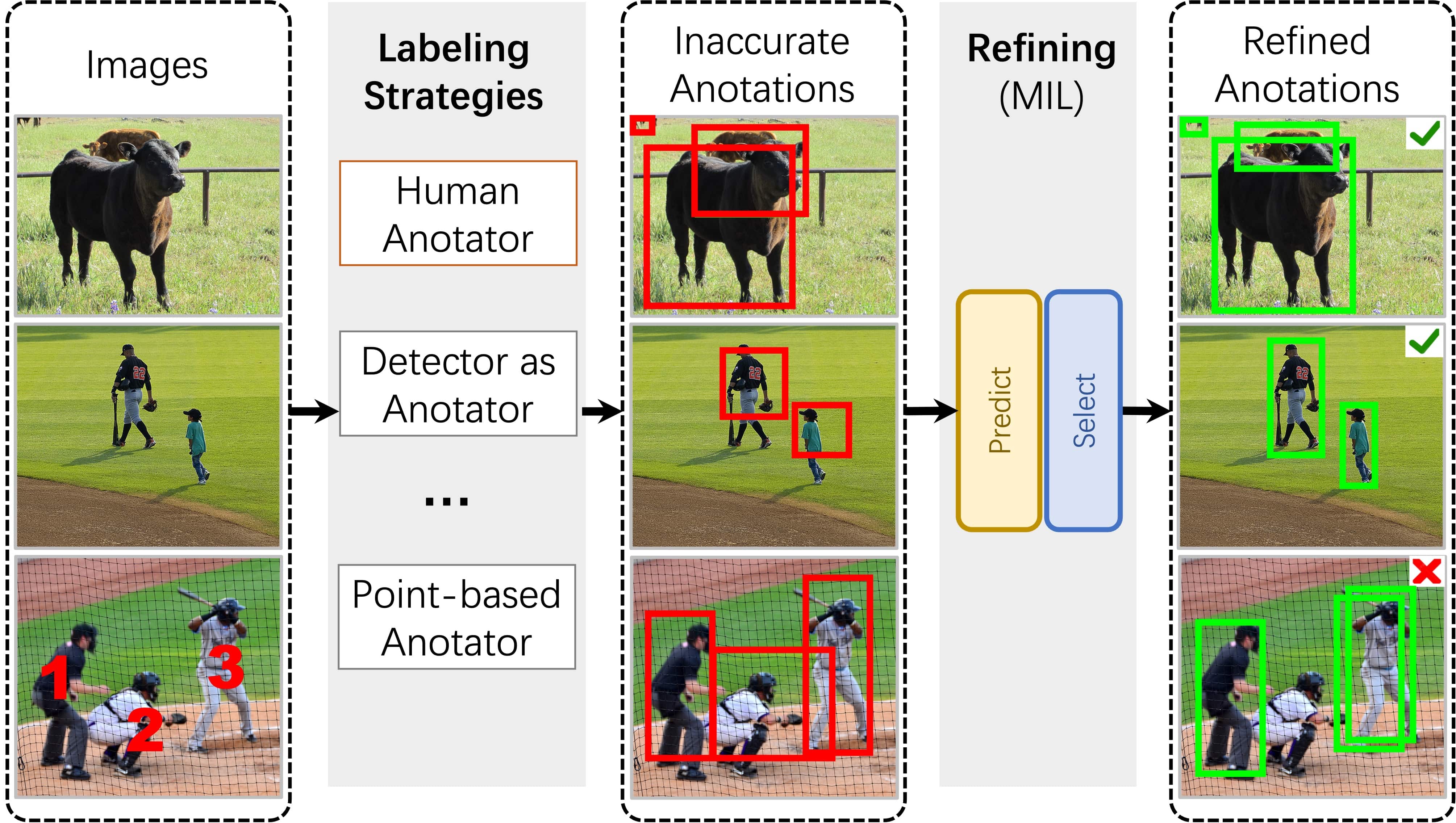}
  }
  \subcaptionbox{Three problems during previous MIL refining.
  Because their selections depend solely on classification,
  (i,ii): the refined box (green box) \textbf{drifts} to another object or makes a \textbf{group prediction} (merging across multiple objects) due to neighbor disturbance.
  Yellow boxes are proposals in the middle person's MIL bag.
  (iii): Local \textbf{part} may be more discriminative than the entire object and will be predicted.
  \label{fig:motivation}}{
  \centering
  \includegraphics[width=\linewidth]{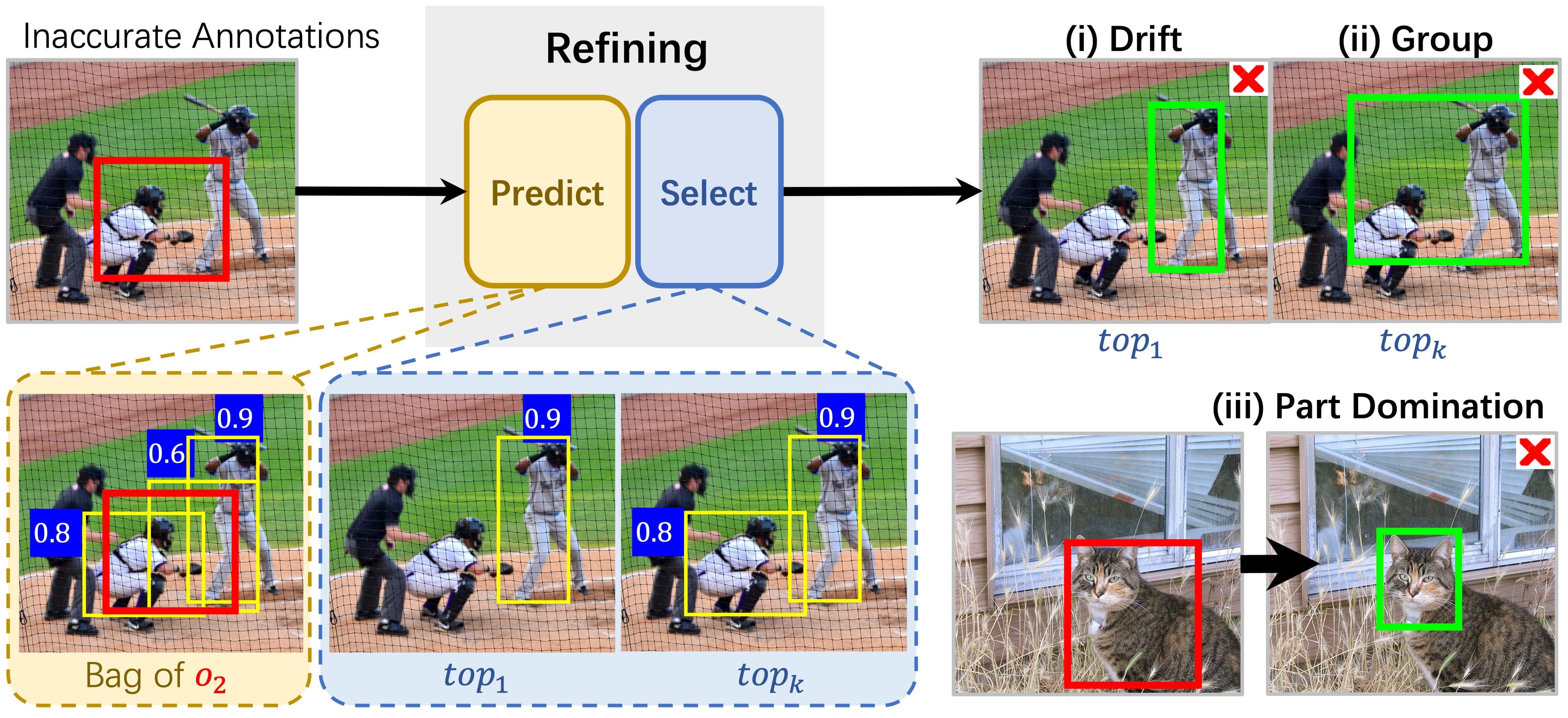}
  }
%\vspace{-10pt}
   \caption{The sources of inaccurate box annotations and three problems caused by previous refinement methods. 
   }
  %\vspace{-20pt}
\end{figure}

Object detection ~\cite{DBLP:MASK-RCNN,FasterRCNN,FreeAnchor,DBLP:retinanet_focalloss, scalematch} relying on large-scale datasets like MS-COCO\cite{coco} has significantly progressed and achieved good performance.
% 1\ 标注方式带来噪声
However, accurate bounding box annotations are expensive and challenging in natural contexts~\cite{oamil}. Especially in many professional scenarios, it is difficult to label accurate annotations without domain knowledge (\eg, agricultural crop observation and medical image processing)~\cite{oamil,DBLP:wsddn}.
% 2\ 目标自身带来噪声
As shown in Fig.~\ref{fig: noisy source} in some complex datasets, the human annotators may also annotate inaccurate bounding boxes due to the inherent ambiguities\cite{KLloss} of objects.
In addition, labelling with a detector or weak signal\cite{p2b} (\eg, point) is much cheaper but brings more inaccuracy. 
Therefore learning robust detectors with inaccurate bounding boxes\cite{oamil,DBLP:IV_NOISY, DBLP:tis_Annotation_Refinement,DBLP:TIP_LABLE_BOX,DBLP:journals/corr/abs-2003-01285} is a practical and meaningful task and has boosted a broad interest.

To use the inaccurate annotations, most related methods~\cite{oamil,p2b} refine the inaccurate annotations as Fig.~\ref{fig: noisy source} shows, and then train a detector head or re-train a detector with the refined box as the new supervision. 
There are two main steps during refining: 1) \textbf{Bag Construction}: For each object, obtaining some proposals around the inaccurate annotated bounding box to form the object-level proposal bag; 2) \textbf{Proposal Selection}: Selecting the top-$k$ proposals with the highest classification confidence from each bag and then weighting average them to obtain the refined box.

During the proposal selection, they usually utilize multiple instance learning (MIL)~\cite{DBLP:MIL} supervised by category information to choose the proposals with high classification confidence from the constructed bags. However, they pay less~\cite{oamil} or no~\cite{p2b} attention to mining spatial information, leading to the following problems as shown in Fig.~\ref{fig:motivation}: 
(1) \textbf{Object Drift}: For each object, some proposals in the constructed bag do not have a high IoU with the original object but with another nearby object. These proposals are not spatially adjacent to the original object but still have high classification confidence, as the rightmost (the yellow box) proposal of $O_2$ shows in the left-bottom corner of Fig.~\ref{fig:motivation}. Only the category confidence is relied on for selecting proposals for $O_2$, and the rightmost proposal will be selected as the refined box. It means the refined box drifts to another object (Fig.~\ref{fig:motivation} (i)), reducing the recall;
(2) \textbf{Group Prediction:} Most works~\cite{DBLP:slv,p2b,CPR} select the top-$k$ proposals by classification confidence and then weight average them as the refined box, causing the group prediction problem, as shown in Fig.~\ref{fig:motivation} (ii);
(3) \textbf{Part Domination}: The detector often focuses on the object's semantic region, which can statistically represent the category (\eg the face). As shown in Fig.~\ref{fig:motivation} (iii), the high classification confidence of the animal is in the discriminant part (the face) rather than the entire object as mentioned by~\cite{DBLP:oicr,Instance-Aware}.

\begin{figure}[t]
  \centering
  \includegraphics[width=\linewidth]{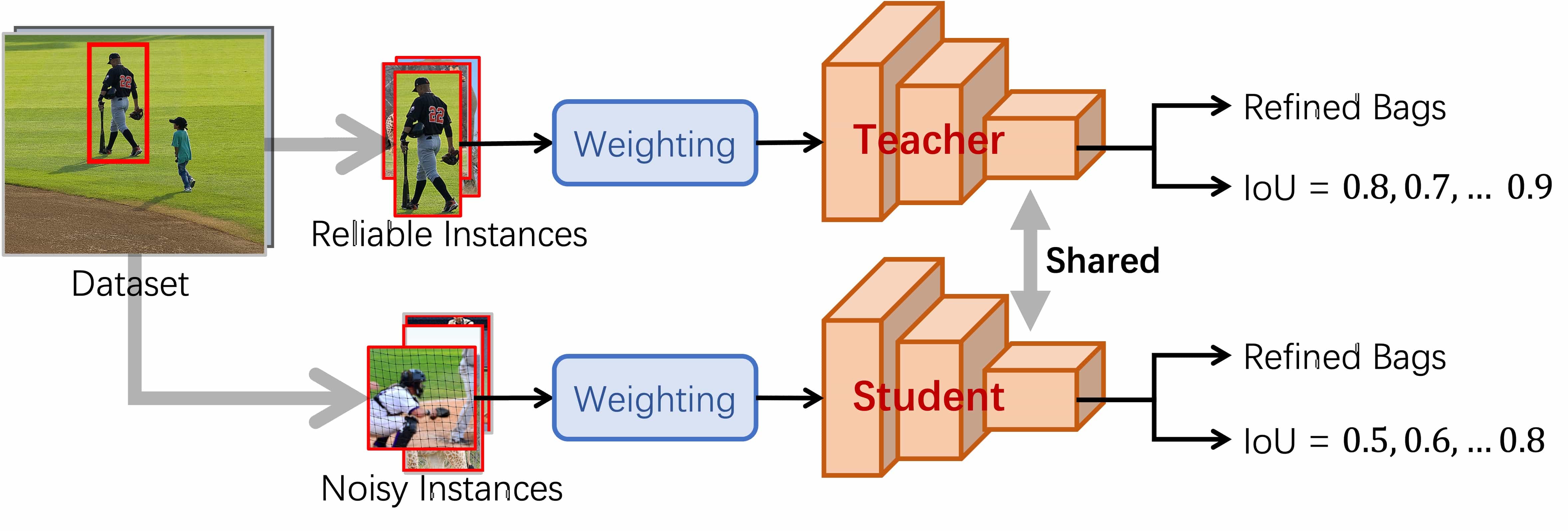}
   \caption{The mechanism of Spatial Self-Distillation. 
   By assigning higher weight, the low-noise annotations can be seen as reliable samples to guide the training of proposals' spatial position and identity learning in SPSD and SISD.}
   \label{fig:distillation}
   %\vspace{-17pt}
\end{figure}

To address these problems, we propose a \textbf{Spatial Self-Distillation} based detector \textbf{(SSD-Det)} to integrate the spatial cues into the bounding box refinement. SSD-Det has two important components: the Spatial Position Self-Distillation \textbf{(SPSD)} module for the bag construction step and the Spatial Identity Self-Distillation \textbf{(SISD)} module for the proposal selection step. 
To construct high-quality proposal bags, SPSD utilizes a neighborhood sampler to generate a balanced and flexible initial proposal bag for each object and then trains a regressor with the supervision of the annotated inaccurate bounding boxes. Finally, high-quality proposal bags are constructed with proposals corrected by the regressor.
The mechanism behind SPSD is that the network learns the spatial information from the reliable samples, \eg those low-noise annotations, in the dataset and then guides the noisy samples to produce high-quality proposals, as shown in Fig.~\ref{fig:distillation}.
In addition, to further combine the category information and the spatial information, an interactive structure is implemented by alternately using SPSD to mine spatial cues and MIL to utilize the category information.
With SPSD and the interactive structure, a high-quality proposal bag can be constructed. 
 Experiments on MS-COCO show that SPSD can significantly improve the mean/max IoU between objects and proposals (about 18/10 points, Fig.~\ref{fig:vis_bag}) in the constructed bag. 
Instead of selecting proposals by classification confidence, we have proposed the SISD module in the proposal selection step. We use it to obtain each proposal's spatial confidence by predicting the IoU with the object and combining the IoU with classification confidence to select the top-$k$ proposals. It is worth mentioning that SISD is an object-related IoU predictor, which means that the predicted IoU may be different for the same proposal that appears in different objects' bags. Accordingly, it guarantees that SISD can better handle object drift and group prediction problems.
Experiments on MS-COCO and VOC datasets verify the effectiveness of our method and bring state-of-the-art performance. The contributions are as follows:

1) We further investigate the inaccurate-box supervised object detection tasks and propose an end-to-end training SSD-Det that combines the spatial and the category information in an interactive fashion.

2) We utilize an SPSD module to generate higher-quality proposals sampling through statistic-guide spatial position distillation, raising the upper bound of the refinement.

3) To add spatial cues to classification confidence, we also introduce an SISD module to select a proposal belonging to the object rather than the category.

4) The performance of our proposed SSD-Det improves the mean average precision ($\rm AP$) of the best previous method (\eg over 10 $\rm AP$ on 40\% noisy MS-COCO) and achieves state-of-the-art under various noise rate box supervision on MS-COCO and VOC datasets.

\section{Related Work}
%\vspace{-3pt}
\subsection{Object Detection}
Classic object detection \cite{fastrcnn,FasterRCNN,DBLP:yolo,DBLP:SSD,DBLP:retinanet_focalloss,DBLP:DETR,DBLP:sparsercnn,scalematch} is supervised by an accurate bounding-box.
One-stage detectors utilize anchors as the sliding-window, such as YOLO~\cite{DBLP:yolo}, SSD~\cite{DBLP:SSD}, and RetinaNet~\cite{DBLP:retinanet_focalloss}. Two-stage detectors mine spatial information to predict proposals (\eg selective search~\cite{DBLP:selectivesearch} in Fast R-CNN~\cite{fastrcnn} or RPN in Faster R-CNN~\cite{FasterRCNN}) and conduct classification and bounding-box regression with filtered proposals sparsely.
Transformer-based (\ie DETR~\cite{DBLP:DETR}, Deformable-DETR~\cite{DBLP:DeformableDETR}, and Swin-Transformer~\cite{DBLP:swin-transformer}) detectors utilize global information for better representation. Sparse R-CNN~\cite{DBLP:sparsercnn} combines a transformer's advantages and CNNs for detection.

\subsection{Weakly-Supervised Object Detection (WSOD)}
WSOD trains object detectors with image tag supervision. Only with the category annotation, the majority of previous methods treat each image as a bag
% (Fig.~\ref{fig:samples} (a)) 
and candidate proposals as instances. They follow the multiple instance learning (MIL) pipeline~\cite{DBLP:wsddn,DBLP:oicr,DBLP:pcl,DBLP:slv,DBLP:MELM}, which highly depends on category information.
However, the MIL loss function leads to a non-convex optimization problem; thus, MIL solutions are usually stuck into the local minima. Context information \cite{ContextLocNet,ts2c}, spatial regularization \cite{DBLP:wsddn,DBLP:wccn,DBLP:MELM}, and optimization strategy \cite{DBLP:oicr,DBLP:MELM,DBLP:pcl} are proposed to address the problems. SPE~\cite{SPE2022} introduces Transformer into WSOD and uses attention to generate proposals. SD-LocNet~\cite{SD-LocNet} tackles the initialized noisy object locations in WSOD and proposes a self-directed localization network to identify the noisy object instances. ~\cite{DBLP:oicr,DBLP:pcl,DBLP:slv} use the pseudo label for classification's iterative refinement. However, we use the pseudo box as a better self-distillation teacher. ~\cite{end2end,Instance-Aware} conduct regression to move the proposals, whereas we conduct regression to distill for better bag construction.
In this work, we also formulate box correction as a MIL problem.

\subsection{Semi-Supervised Object Detection (SSOD)}
Semi-supervised learning in object detection can be roughly categorized into two groups: consistency based~\cite{DBLP:conf/nips/JeongLKK19,DBLP:conf/wacv/TangRWXX21} and pseudo label based~\cite{DBLP:conf/eccv/LiHQWG20,DBLP:conf/cvpr/RadosavovicDGGH18,DBLP:journals/corr/abs-2005-04757,DBLP:conf/cvpr/WangYZ0L18,DBLP:conf/nips/ZophGLCLC020,soft-teacher}.
\cite{soft-teacher} presents an end-to-end SSOD approach with two simple techniques
named soft teacher and box jittering to facilitate the efficient leverage of the teacher model.
Both Soft Teacher~\cite{soft-teacher} and SSD-Det obtain pseudo-labels from candidate boxes, adopting distillation, box jitter and classification scores weighting policy. However, they are different:
1) SSOD selects candidate boxes from the teacher model's detection results, while SSD-Det generates them with a generative approach, due to no any accurate supervised data for ensuring high-quality boxes in the detection results.
2) ~\cite{soft-teacher} is based on FixMatch~\cite{FixMatch} and requires maintaining two networks for teacher-student distillation structure. In contrast, our approach only needs a multi-head detector with a shared backbone for self-distillation.
3) Box Jitter: ~\cite{soft-teacher} calculates box variance from jittering for result selection, while we aim to generate candidate boxes that combine SISD-predicted IoU and classification scores, selecting boxes closer to the ground truth.
4) We use classification scores as weights for the next stage's loss, while ~\cite{soft-teacher} employs them as a criterion for selecting reliable samples.

%\vspace{-2pt}
\subsection{Learning with Noisy Annotations}
Training CNNs under noisy labels has been an active research area. Previous research focuses on the classification task, and develops various techniques to deal with noisy labels, such as sample selection ~\cite{DBLP:co-teaching,DBLP:Mentornet} for training, label correction~\cite{DBLP:SELFIE,DBLP:Dimensionality-Driven}, and robust loss functions~\cite{DBLP:GCE,DBLP:Robust_Loss} against noisy labels. Recently, many efforts~\cite{DBLP:IV_NOISY,DBLP:journals/corr/abs-2003-01285, DBLP:tis_Annotation_Refinement,DBLP:TIP_LABLE_BOX,p2b,oamil} have been devoted to the object detection task. On the one hand, Simon \etal \cite{DBLP:IV_NOISY} first investigates the impact of different types of label noise on object detection. They propose a per-object co-teaching method to alleviate the effect of noisy labels. On the other hand, \cite{DBLP:TIP_LABLE_BOX} proposes a meta-learning framework for noisy annotations consisting of noisy category labels and bounding boxes. \cite{p2b,oamil} utilize object-level MIL to refine the inaccurate box. OA-MIL~\cite{oamil} constructs proposal bags through label assignment 
% (shown in Fig.~\ref{fig:samples} (b)) 
in a discriminant style. P2BNet~\cite{p2b} originally conducts point-supervised object detection tasks. However, it can be seen as the box correction in its refinement stage. It uses hand-craft anchors
% (given in Fig.~\ref{fig:samples} (c)) 
to generate proposal bags. Our method inherits the generative style of P2BNet and conducts spatial distillation to mine spatial information.

\subsection{Knowledge Distillation}
Knowledge distillation (KD) \cite{hinton2015distilling} aims to learn compact and efficient student models guided by excellent teacher networks.
It is first applied to object detection in \cite{DBLP:kdod},  in which hint learning and KD are both used for multi-class object detection.
Recently, many efforts \cite{DBLP:conf/cvpr/LiJY17,DBLP:finegrained,DBLP:conf/cvpr/DaiJWBW0Z21,DBLP:conf/cvpr/Guo00W0X021,DBLP:conf/cvpr/YangLJGYZY22} aim to mimic the feature.
\cite{DBLP:conf/cvpr/ZhengYWRZHC22} shows that localization knowledge is more important and proposed a localization distillation method.
We also transfer the spatial knowledge from reliable labeled instances to correct inaccurate bounding boxes (shown in Fig.~\ref{fig:distillation}) in a self-distillation manner.

%\vspace{-3pt}
\section{Methodology}
\begin{figure*}[t]
  \centering
  \includegraphics[width=\linewidth]{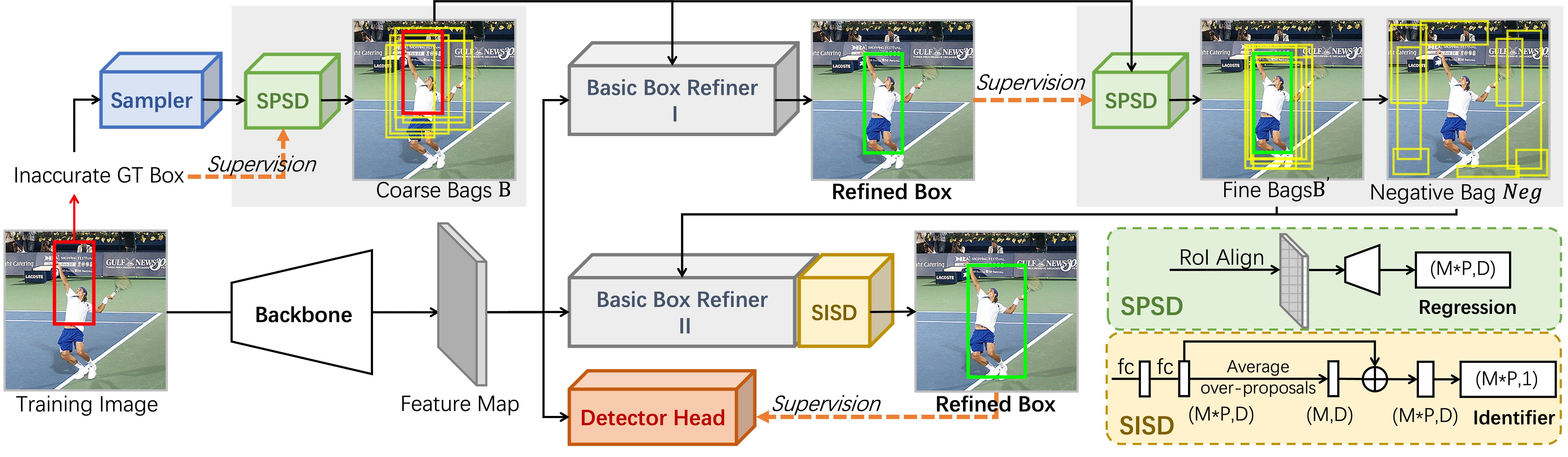}
  %\vspace{-10pt}
  \caption{The framework of SSD-Det. It contains basic box refiner, SPSD module, SISD module and a detector head. Neighborhood sampler is adopted around the inaccurate annotation. Then, SPSD module generates better proposal bags which are fed into basic box refiner for MIL training. The selected proposals are average weighted as the refined box and supervise the next SPSD training. Meanwhile, the SISD module predicts the IoU between proposals and the object, and the estimated IoU is multiplied by classification score for better proposal selection to generate the refined box. SPSD shares backbone with the detector.}
   \label{fig:framework}
   %\vspace{-10pt}
\end{figure*}

%\vspace{-3pt}
This work aims to learn a robust detector with inaccurate bounding boxes. 
Instead of training a detector with the original inaccurate bounding box, we follow most related works \cite{p2b, oamil} that design a branch to refine the inaccurate bounding box and then train the detector head or detector with the refined bounding box.
The most important part is how to design a refining policy.
We first design a two-stage basic box refiner (gray region in Fig.~\ref{fig:framework}) as a naive solution that modified from \cite{p2b}. Then, SPSD and SISD are proposed and added to further mine the spatial cues for box refinement, yielding SSD-Det.
Therefore, the overall loss function is formulated as:
%\vspace{-3pt}
\begin{equation}\small
    \begin{aligned}
      \mathcal{L}=
      % \alpha_1 \cdot 
      \mathcal{L}_{Basic}
      + \alpha_1 \cdot \mathcal{L}_{SPSD} 
      + \alpha_2 \cdot \mathcal{L}_{SISD} + \alpha_3 \cdot \mathcal{L}_{Det}
    \end{aligned}
\end{equation}
\noindent where $\alpha_1, \alpha_2$ and $\alpha_3$ are set as 0.25, 0.25 and 4 respectively. $\mathcal{L}_{Det}$ denotes the loss of detector or detection head.
During inference, only the detector or the detection head is used.

\subsection{Basic Box Refiner\label{sec: basic}}
%\vspace{-2pt}

Motivated by \cite{oamil} and WSOD \cite{DBLP:wsddn}, we design the basic box refiner (detailed structure figure is in supplementary) that leverages classification confidence to refine the inaccurate box annotation. Then the refine annotation is used to train a detection head or detector. 
Following ~\cite{p2b}, we design a two-stream structure as a MIL classifier to select the best proposal for box refinement.

Giving an image with inaccurate box annotation, for each object, $\mathcal{B}$ is a bag of proposals (bounding boxes) that are generated around its inaccurate annotation by a sampler policy (\eg, selective search\cite{DBLP:selectivesearch}, edge box\cite{DBLP:EdgeBox}, neighborhood sampler in Sec.~\ref{sec: neighborhood sampler}). Meanwhile a feature map is extracted with a backbone network. And then through $7 \times 7$ RoIAlign~\cite{DBLP:MASK-RCNN} and two fully connected (fc) layers, features of proposal in $\mathcal{B}$ are extracted and denote as $\mathbf{F}$. 
The basic box refiner takes proposal bag $\mathcal{B} \in \mathbb{R}^{P \times 4}$ and features $\mathbf{F} \in \mathbb{R}^{P \times D}$ as inputs, where $P$, $D$ are denoted as the number of proposals in $\mathcal{B}$, feature dimension respectively.

% MIL: cls
Following ~\cite{p2b} and ~\cite{DBLP:wsddn}, as Eq.~\ref{Eq:MIL score} described, we apply the classification branch $f_{cls}$ to $\mathbf{F}$ yields $\mathbf{O}^{cls}$, which is then passed through the $softmax$ function over classification dimension $K$ to obtain the score $\mathbf{S}^{cls}\in \mathbb{R}^{P \times K}$, where $K$ represents the number of instance categories. 
% MIL: ins
Likewise, instance selection branch $f_{cls}$ is applied to $\mathbf{F}$ to yield $\mathbf{O}^{ins}$, and instance score $\mathbf{S}^{ins}$ is obtained through $softmax$ function over $P$ proposals.
%  MIL: proposal and bag score
The proposal score $\mathbf{S}$ is obtained by computing the Hadamard product of the classification score and the instance score. The bag score $\widehat{\mathbf{S}}$ is obtained by the summating of the $P$ proposal boxes' proposal scores.
%\vspace{-10pt}
\begin{equation}\footnotesize
\begin{aligned}
& \mathbf{O}^{cls} = f_{cls}(\mathbf{F}) \in \mathbb{R}^{P \times K};
[\mathbf{S}^{cls}]_{pk} = e^{[\mathbf{O}^{cls}]_{pk}}\big/\sum_{k=1}^{K} e^{[\mathbf{O}^{cls}]_{pk}}. \\
& \mathbf{O}^{ins} = f_{ins}(\mathbf{F}) \in \mathbb{R}^{P \times K}; 
[\mathbf{S}^{ins}]_{pk} = e^{[\mathbf{O}^{ins}]_{pk}}\big/\sum_{p=1}^P e^{[\mathbf{O}^{ins}]_{pk}}.  \\
& \mathbf{S}=\mathbf{S}^{cls} \odot \mathbf{S}^{ins} \in \mathbb{R}^{P \times K};
\widehat{\mathbf{S}}= \sum\limits_{p=1}^{P} [\mathbf{S}]_p \in \mathbb{R}^{K}.
\label{Eq:MIL score}
\end{aligned} 
%\vspace{-5pt}
\end{equation}
where $[\cdot]_{pk}$ is the value at row $p$ and column $k$ in the matrix. 

% loss basicI
The basic box refiner has two similar stages. The loss of stage I(termed $\mathcal{L}_{I}$) adopt the MIL paradigm with the form of cross-entropy (CE) loss, defined as: 
%\vspace{-7pt}
\begin{equation}\footnotesize
\begin{aligned}
\mathcal{L}_{I} = CE(\widehat{\mathbf{S}}, \mathbf{c}) &=-\sum\limits_{k=1}^{K} \mathbf{c}_k \log(\widehat{\mathbf{S}}_k) + (1-\mathbf{c}_k)  \log(1-\widehat{\mathbf{S}}_k)
\label{Eq:bce loss}
\end{aligned} 
%\vspace{-3pt}
\end{equation}where $\mathbf{c} \in \{0, 1\}^{K}$ is the one-hot category label. And each object's proposals with the top-$k$ highest proposal score $\mathbf{S}$ are weighted to obtain the refined box.

The stage II takes the refined box of stage I as input and performs fine refining with a similar structure as stage I.
Differently, the focal loss is adopted in stage II instead of cross entropy loss. In order to cooperate with focal loss, the classification branch uses the $sigmoid$ $\sigma(x)$ instead of $softmax$ function and we sample some negative samples $\mathcal{N}$ to further suppress the background.
With the bag score ${\widehat{S}}$ and the negative sample scores $\mathbf{S}^{cls}_{neg}$, the loss is:
%%\vspace{-0.2cm}
%\vspace{-9pt}
\begin{equation}\small
\begin{aligned}
\setlength\abovedisplayskip{0pt}%shrink space
\setlength\belowdisplayskip{1pt}
\mathcal{L}_{II}  &= \left< \mathbf{c}^{\mathrm{T}}, \mathbf{\widehat{S}}^* \right> \cdot {\rm FL}(\mathbf{\widehat{S}}, \mathbf{c}) + \sum\limits_{\mathcal{N}} \beta  \cdot {\rm FL}(\mathbf{S}^{cls}_{neg},c_{neg})\big\}
\label{Eq:L_{bag}}
\end{aligned}
%\vspace{-9pt}
\end{equation}
\noindent where ${\rm FL}$ is the focal loss~\cite{DBLP:retinanet_focalloss},
$\mathbf{\widehat{S}}^*_j$ represents the bag score predicted by stage I. ${\small{\left< \mathbf{c}^{\mathrm{T}}_j, \mathbf{\widehat{S}}^*_j \right>}}$ represents the inner product of the two vectors, meaning the predicted bag score of the ground-truth category. $\beta$ is the average of ${\small{\left< \mathbf{c}^{\mathrm{T}}_j, \mathbf{\widehat{S}}^*_j \right>}}$. 
% The proposal score $\mathbf{S}$
They are used to weight each object's $\rm FL$ for stable training.
The overall loss function of the basic refiner here is:
%\vspace{-8pt}
\begin{equation}\small
\mathcal{L}_{Basic}=\mathcal{L}_{I}+\alpha_{II}\cdot\mathcal{L}_{II}
%\vspace{-6pt}
\end{equation}
\noindent where $\alpha_{II}$ are the loss weights of the two stages.

During training, the refined box of stage II is used as supervision for a detection head or detector. After training, the basic box refiner will be removed, leaving a well-trained detection head or detector. In this way, we can train a detector under inaccurate annotations.  

\subsection{Spatial Position Self-Distillation (SPSD)}

Like most MIL paradigm methods, basic box refiner has two main components: bag construction and proposal selection. And the main idea is to use classification information to guide the refining. In this paper, we add spatial information to improve refining. Specifically, SPSD is proposed to use spatial information to enhance bag construction.

Bag construction aims to obtain proposals for each object, while proposal selection is to select the proposals from the object bag. Then, the refined box is averaged over the selected proposals. Therefore, the quality of the proposals in constructed bag determines the upbound of refining. The bag construction can be implemented in a variety of ways.
In this paper, the basic box refiner adopts a naive neighborhood sampler for bag construction. basic box refiner adopts a naive neighborhood sampler for bag construction.

\textbf{Neighborhood Sampler. \label{sec: neighborhood sampler}} Proposals around the inaccurate box are sampled to construct an object bag. 
For each inaccurate box $b^*=(b^*_x,b^*_y,b^*_w,b^*_h)$, its scale and aspect ratio with $s$ and $v$ are adjusted and its positions $o_x, o_y$ are jittered to obtain the diverse proposal $b=(b_x, b_y, b_w, b_h)$:
\begin{equation}\small
\setlength\abovedisplayskip{2pt}%shrink space
\setlength\belowdisplayskip{0pt}
\begin{aligned}
% b_w = v \cdot s \cdot b^*_w, \quad b_h = \frac{1}{v} \cdot s \cdot b^*_h, \\
b_w = v \cdot s \cdot b^*_w, \quad b_h = 1/v \cdot s \cdot b^*_h, \\
b_x = b^*_x + b_w \cdot o_x, \quad b_y = b^*_y + b_h \cdot o_y.
\setlength\abovedisplayskip{0pt}
\end{aligned}
\label{Eq: PRB sampling}
\end{equation}

These proposals $b$ are used to construct the positive proposal bag $\mathcal{B}$ to train the MIL classifier.
Thanks to the hand-craft sampling way, the number of proposals in different objects' proposal bags is controllable and balanced. However, the hand-craft neighborhood sampler strategy is difficult to set hyper-parameters, and the sampling space is discrete. For example, when the jitter region is small, the optimization space of refining is limited, while when it is large, more background will be introduced. 
Hence, we propose the SPSD module to mine spatial information for higher-quality proposal bag construction.

\textbf{Statistically Guided Adaptive Sampling.} 
Instead of simply using a neighborhood sampler, we adopt a statistically guided adaptive sampling by adding SPSD modules into the basic box refiner.
Taking the constructed proposal bag $\mathcal{B}$ of hand-craft neighborhood sampler as input, the RoI features of proposals in $\mathcal{B}$ are extracted and fed into the two shared fc layers to obtain $\mathbf{F}$. Then a regression fc layer $f_{dis}$, supervised by the inaccurate annotated bounding box $b^*$, is introduced to predict the adaptive proposal bag $\mathcal{B}^{dis}=f_{dis}(\mathbf{F}) \in \mathbb{R}^{P \times 4}$, in which the proposals are closer to the object. Later, $\mathcal{B}^{dis}$ as the constructed proposal bag is fed into stage I of basic box refiner.
In order to combine category and spatial information, we implement an interactive structure by alternately using SPSD to mine the spatial cues and using MIL in basic box refiner to utilize the category information. 
Specifically, the refined bounding box $\hat{b^*}$ of stage I that selected by the classification confidence is used to supervision of a new SPSD module for stage II. 
 Similar as stage I, The new SPSD takes proposal bag $\mathcal{B}^{dis}$ of hand-craft neighborhood sampler as input. Through the RoI align and the two shared fc layers, the feature $\hat{\mathbf{F}}$ is extracted. An extra fc layer $\hat{f_{dis}}$ is then utilized to conduct further regression. Different with stage I, the obtained $\hat{\mathcal{B}^{dis}}$ is supervised by the refined $\hat{b^*}$.
The loss function of the spatial distillation for adaptive sampling can be defined as $\mathcal{L}_{SPSD}$ in Eq.~\ref{Eq: l_distill}.
%\vspace{-9pt}
\begin{equation}\small
\mathcal{L}_{SPSD}= \frac{1}{P}\big\{\sum_{p=1}^P\mathbf{L_1}([\mathcal{B}^{dis}]_p,b^*) +\sum_{p=1}^P\mathbf{L_1}([\hat{\mathcal{B}^{dis}}]_p,\hat{b^*})\big\}
\label{Eq: l_distill}
%\vspace{-9pt}
\end{equation}
\noindent where the $\mathbf{L_1}$ is the L1 loss function for loose restrictions.

The idea behind SPSD is that the dataset with inaccurate annotation still has many reliable, high-quality boxes and inaccurate boxes. Supervised by the high-quality boxes statistically, the network can guide those proposals sampled around the inaccurate bounding box to regress to the ground truth.
With the self-distillation mechanism, SPSD learns the semantic-spatial correspondence knowledge from the reliable samples in the dataset and then propagates the knowledge to produce high-quality proposals.

\textbf{Adaptive Negative Sampling.}
Negative samples are introduced in Stage II to better suppress the background. With the sampled $\hat{\mathcal{B}^{dis}}$, we can adaptively sample the negative samples with a small IoU (set smaller than 0.3 by default) with all positive proposals in all bags, to compose the negative sample set $\mathcal{N}$ for stage II.

\subsection{Spatial Identity Self-Distillation (SISD)}
The basic box refiner selects the proposals only depending on classification confidence during proposal selection. To select the proposal which has high classification confidence and is also spatially close to the object from the bag, we propose a SISD module to predict the IoU between proposals and their corresponding object. Afterwards, through the combination between the IoU and the classification confidence, top-$k$ proposals are selected. 
In SISD, we design an Object Relevance Enhancement (ORE) module to distinguish different objects' features with the same RoI region. And an identity predictor is designed to predict each proposal's IoU with the object.

\begin{table*}[tb!]
    \centering
    \resizebox{\textwidth}{!}{
    \begin{tabular}{l|c|cccccc|cccccc}
    \specialrule{0.13em}{0pt}{0pt}
    \multirow{2}{*}{Method} & \multirow{2}{*}{Backbone} & \multicolumn{6}{c|}{20\% Box Noise Level}& \multicolumn{6}{c}{40\% Box Noise Level} \\
    
    &&$AP$&$AP_{50}$&$AP_{75}$&$AP^s$&$AP^m$ & $AP^l$&$AP$&$AP_{50}$&$AP_{75}$&$AP^s$&$AP^m$ & $AP^l$ \\
    \hline
    \hline
    \multicolumn{14}{c}{$Val$ $Set$}\\
     \hline
      Clean-FasterRCNN~\cite{FasterRCNN} & ResNet-50&37.9 &58.1 &40.9&21.6 &41.6 &48.7&37.9 &58.1 &40.9&21.6 &41.6 &48.7 \\
      Clean-FasterRCNN~\cite{FasterRCNN} & ResNet-101&39.4 &60.1& 43.1& 22.4& 43.7& 51.1&39.4 &60.1& 43.1& 22.4& 43.7& 51.1 \\
      Clean-Retinanet~\cite{DBLP:retinanet_focalloss} & ResNet-50 &36.7 & 56.1 & 39.0 & 21.6 & 40.4 &47.4 & 36.7 & 56.1 & 39.0 & 21.6 & 40.4 &47.4 \\
      \hline
        % Clean-SparseRCNN &ResNet-50 & \\
        % Clean-De-DETR &ResNet-50 \\
      Noisy-FasterRCNN~\cite{FasterRCNN} &ResNet-50&30.4 &54.3 &31.4&17.4 &33.9& 38.7&10.3& 28.9&3.3 &5.7& 11.8& 15.1 \\
      % Noisy-SparseRCNN &ResNet-50 & \\
      Noisy-Retinanet~\cite{DBLP:retinanet_focalloss} & ResNet-50 & 30.0 &53.1 & 30.8&17.9&33.7&38.2 & 13.3 &33.6 & 5.7&8.4&15.9&18.0 \\
      FreeAnchor\cite{FreeAnchor} &ResNet-50&28.6& 53.1&28.5& 16.6& 32.2& 37.0&10.4& 28.9&3.3 & 5.8& 12.1& 14.9 \\
      Co-teaching\cite{DBLP:co-teaching} &ResNet-50&30.5& 54.9&30.5& 17.3& 34.0& 39.1&11.5& 31.4&4.2 &6.4& 13.1& 16.4\\
      SD-LocNet\cite{SD-LocNet} &ResNet-50&30.0& 54.5&30.3& 17.5& 33.6& 38.7& 11.3& 30.3&4.3 & 6.0& 12.7& 16.6 \\
      KL loss\cite{KLloss} &ResNet-50&31.0& 54.3&32.4& 18.0& 34.9& 39.5&12.1& 36.7&3.7& 6.2& 13.0& 17.4 \\

      OA-MIL\cite{oamil} &ResNet-50&32.1& 55.3 &33.2& 18.1& 35.8& 41.6& 18.6& 42.6&12.9& 9.2& 19.9& 26.5 \\
    %   Clean-FasterRCNN &&&&&& \\
      \hline
      SSD-Det & ResNet-50    & 33.6     & 57.3      &     35.3  & 19.5      & 37.2      &     43.3  & 27.6     & 53.9      &    26.0   & 16.0      & 31.0     & 34.9 \\
    %   SSD-Det & ResNet-101 &&&&& \\
    %   SSD-Det+ & ResNet-50 &&&&& \\
      SSD-Det & ResNet-101 &34.3& 57.6 & 36.7 & 19.1&38.1 & 44.3& 28.4 &  54.3 & 27.2 & 16.5 & 31.9 & 36.4 \\
      SSD-Det+FR & ResNet-50 & 34.4   &   57.3   &  36.8 &  20.0    &    38.2   &   44.0 &29.3   & 54.8      & 29.0    &    17.1  & 32.9    &  36.9\\
    %   SSD-Det* & ResNet-101 &&&&& \\
    %   SSD-Det*+ & ResNet-50 &&&&& \\
      SSD-Det+FR & ResNet-101 &36.2&59.1& 39.2&  20.9&40.2 &47.1 &30.6 &56.7 &30.7 &18.1 &34.5&39.0\\
      % SSD-Det+SR &ResNet-50 &40.1&63.0&44.3&26.2&43.5&52.0&34.3&60.2&36.4&22.4&37.5&43.7\\
      % SSD-Det+De-DETR &ResNet-50&42.1&64.5&46.4&26.8&45.7&54.7&35.0&60.7&37.4&23.6&38.1&44.4\\
      \hline
      \hline
      \multicolumn{14}{c}{$Test$ $Set$}\\
      \hline
      Clean-FasterRCNN~\cite{FasterRCNN} & ResNet-50& 37.7&58.7&40.8&21.7&40.6&46.7&37.7&58.7&40.8&21.7&40.6&46.7\\
      \hline
      Noisy-FasterRCNN~\cite{FasterRCNN} &ResNet-50&30.7& 54.9&31.3&18.0&33.7&37.7&10.4&29.0&3.3&6.0&11.3&14.6\\
      OA-MIL\cite{oamil} &ResNet-50& 32.3&55.8&33.7&18.5&35.0&40.2&18.5&42.3&12.8&9.3&19.1&25.1\\
      \hline
      SSD-Det & ResNet-50    & 33.5&57.3&35.5&19.1&36.0&41.9&28.0&54.1&26.5&16.5&30.0&34.5\\
      SSD-Det+FR & ResNet-50 & 34.7&57.9&37.2&20.0&37.7&42.7&29.7&55.6&29.3&17.5&32.4&36.2\\
    \specialrule{0.13em}{0pt}{0pt}
    \end{tabular}}
    %\vspace{-3pt}
    \caption{Performance comparison on COCO% validation set, where * meanss Re-Train. 
    . FR is Faster R-CNN. *-FR refers to a retrained Faster R-CNN (R50+FPN) using refined annotations from SSD-Det for improved performance.%, SR means Sparse R-CNN and De-DETR means Deformable DETR
    . Clean-* and Noisy-* means original and noisy annotation.}
    \label{tab:coco-table}
    %\vspace{-10pt}
\end{table*}

\textbf{Object Relevance Enhancement (ORE)}. ORE enhances object-relevant features, making SISD an object-relevant IoU predictor. ORE allows the predicted IoU to be different for the same proposal in other objects' bags.
In addition, we integrate the feature of the bag's corresponding object into the proposal feature, making the feature of different bags' proposals distinct. That is the so-called ORE.
For a proposal bag $\mathcal{B}$, the feature $\mathbf{F}$ is obtained through the RoI align and two fc layers. It is worth mentioning that the two fcs do not share the parameters with those in the refiner since the optimization goals are contradictory.
To represent the feature of the bag's corresponding object, $\mathbf{F}^+ \in \mathbb{R}^{1 \times D}$ is calculated by averaging features of $P$ proposals in proposal bag $\mathcal{B}$. The object feature $\mathbf{F}^+$ is broadcast into $\mathbb{R}^{P \times D}$, and then added to the proposal features to obtain the object-relevant features $\mathbf{F}^*=\mathbf{F}+\mathbf{F}^+$. 

\textbf{Spatial Identity Prediction.} By a following identity fc layer, $U \in \mathbb{R}^{P \times 1}$ is predicted. 
The pseudo label $T \in (0,1)$ is IoU between proposals in $\mathcal{B}$ and the merged box $\hat{b^*}$ of stage I.
For better optimization, the linear normalized $T'=(T-0.5)/0.5 \in (-1,1)$ is utilized as supervision. The object function of the identity predictor is identified in:
% %\vspace{-10pt}
\begin{equation}\small
    \begin{aligned}
    \mathcal{L}_{SISD}= smooth_{L1}(U,T')
    \end{aligned}
    \label{Iou loss}
    %\vspace{-4pt}
\end{equation}
\noindent where the $smooth_{L1}$ represents the smooth L1 loss. The predicted spatial confidence $U'$ is obtained by normalizing the $U$. Finally, $S^*=U' \cdot S $ is used to select the top-$k$ proposals for merging as the refined boxes.

\section{Experiment}
\subsection{Experimental Settings}

\begin{table}[tb!]
    \centering
    % \scriptsize
    \resizebox{0.47\textwidth}{!}{
    \begin{tabular}{l|c|cccc}
    \specialrule{0.13em}{0pt}{0pt}
    \multirow{2}{*}{Method} & \multirow{2}{*}{Backbone} &
    \multicolumn{4}{c}{Box Noise Level} \\
    
    &&10\%&20\%&30\%&40\%  \\
      \hline
      Clean-FasterRCNN~\cite{FasterRCNN} & ResNet-50&\multicolumn{4}{c}{77.2 $for$ $clean$} \\
    %   Clean-FasterRCNN~\cite{FasterRCNN} & ResNet-101&\multicolumn{4}{c}{79.40 $for$ $clean$}\\
      Clean-RetinaNet~\cite{DBLP:retinanet_focalloss}&ResNet-50& \multicolumn{4}{c}{73.5 $for$ $clean$}\\
      \hline
      Noisy-FasterRCNN~\cite{FasterRCNN} &ResNet-50&76.3&71.2&60.1&42.5\\
      Noisy-RetinaNet~\cite{DBLP:retinanet_focalloss}&ResNet-50& 71.5& 67.6&57.9&45.0\\
      KL loss\cite{KLloss} &ResNet-50&75.8& 72.7& 64.6& 48.6\\
      Co-teaching\cite{DBLP:co-teaching} &ResNet-50&75.4& 70.6& 60.9& 43.7\\
      SD-LocNet\cite{SD-LocNet} &ResNet-50&75.7& 71.5& 60.8& 43.9\\
      FreeAnchor\cite{FreeAnchor} &ResNet-50&73.0& 67.5& 56.2& 41.6\\
      OA-MIL\cite{oamil} &ResNet-50&77.4& 74.3& 70.6& 63.8\\
    %   OA-MIL*\cite{oamil} &ResNet-50&77.40& 74.30& 70.60& 63.80\\
      \hline
      SSD-Det & ResNet-50  & 77.1& 74.8& 71.5& 66.9\\
    \specialrule{0.13em}{0pt}{0pt}
    \end{tabular}}
    \caption{Performance comparison on the VOC 2007 test set. The evaluation metric is $AP_{50}$. The Clean-* and Noisy-* means original annotation and noisy annotation.}
    \label{tab:voc-table}
    %\vspace{-15pt}
\end{table}

\begin{table*}[htbp]
  \centering
    \footnotesize
    \setlength\tabcolsep{2pt}
  \resizebox{\textwidth}{!}{
    \begin{tabular}{c|ccc|cccccc|cc|cccccc|cc}
    \specialrule{0.13em}{0pt}{0pt}
    \multirow{2}[2]{*}{2-Ref} 
    % & \multirow{2}[2]{*}{B.S.} 
    & \multirow{2}[2]{*}{SPSD} & \multirow{2}[2]{*}{SISD} & \multirow{2}[2]{*}{Re-Train} & \multicolumn{8}{c|}{20\% Box Noise Level}                     & \multicolumn{8}{c}{40\% Box Noise Level} \\
          &       &       &       & $AP$    & $AP_{50}$  & $AP_{75}$  & $AP^s$   & $AP^m$   & \multicolumn{1}{c}{$AP^l$} &$AP^{test}$    & $AP_{75}^{test}$  & $AP$    & {$AP_{50}$} & {$AP_{75}$} & \multicolumn{1}{c}{$AP^s$} & \multicolumn{1}{c}{$AP^m$} & \multicolumn{1}{c}{$AP^l$} & $AP^{test}$    & {$AP_{75}^{test}$}\\
    \hline
          &        &       &       &  30.0     & 57.1      & 29.0      &16.9       &    33.1   & 39.8    &-&-  & 22.8 & 51.1 &    16.1   & 13.3      & 25.0      &30.4  &-&-\\
    \checkmark         &       &       &       & 31.2 & 56.7 &  31.6     &  17.8     &   34.5    &   41.0   &31.4&32.0 & 24.6 & 52.0 &  20.1     &  14.3     & 28.2      & 31.9&25.0&20.5 \\
    \checkmark     & \checkmark     &       &       & 33.0 & 56.9 &   34.8    &  18.7     & 35.5      &    42.2  &33.1&34.8 & 27.2 & 53.7 & 24.7      &  15.9     & 30.3      & \textbf{35.2}&27.6&25.6 \\
    \checkmark     & \checkmark     & \checkmark &&   \textbf{33.6}     & \textbf{57.3}      &     \textbf{35.3}  & \textbf{19.5}      & \textbf{37.2}      &     \textbf{43.3} &\textbf{33.5}& \textbf{35.5}&  \textbf{27.6}      & \textbf{53.9}      &    \textbf{26.0}   & \textbf{16.0}      & \textbf{31.0}      & 34.9 &\textbf{28.0}&\textbf{26.5}\\
    \hline
    \checkmark      &       &       & \checkmark     & 31.8     & 56.8     & 33.1      & 18.4      &35.7       & 40.8 &32.3&33.7     & 26.5 & 54.0 &  23.3    &     15.7  &    30.3   & 33.8&26.8&23.3 \\
    \checkmark    & \checkmark     &       & \checkmark     &     34.1  & \textbf{57.6}      & 36.4      &     19.0  &    37.7   & 43.8      &34.3   &36.6& 29.0   &\textbf{55.1}       & 27.8      &    17.0   &32.5       & 36.7&29.3&28.4  \\
    \checkmark    & \checkmark     & \checkmark     & \checkmark     &   \textbf{34.4}   &   57.3   &  \textbf{36.8}  &  \textbf{20.0}    &    \textbf{38.2}   &   \textbf{44.0}  &\textbf{34.7 }&  \textbf{37.2}&   \textbf{29.3}    & 54.8      &  \textbf{29.0}     &    \textbf{17.1}   & \textbf{32.9}      &  \textbf{36.9}&\textbf{29.7}&\textbf{29.3}\\
    \specialrule{0.13em}{0pt}{0pt}
    \end{tabular}%
    }
    %\vspace{-10pt}
    \caption{Modules ablation of SPSD, SISD and Re-Train on MS-COCO validation set (without) and test set (with test). The Re-Train means we generate the pseudo label by SSD-Det and re-train a Faster R-CNN detector.}
    \label{tab:ablation on modules}
    %\vspace{-5pt}
\end{table*}%

\textbf{Datasets and Evaluation Metrics.}
For experimental comparisons, two publicly available datasets are used for object detection with inaccurate bounding boxes: MS-COCO \cite{coco} and PASCAL VOC 2007 \cite{DBLP:VOC}. \textbf{MS-COCO} (2017 version) has 118k training and 5k validation images with 80 common object categories. 
\textbf{PASCAL VOC} 2007 is one of the most popular benchmarks in generic object detection with 20 classes.
 
\textbf{Evaluation Metric.} We use mean average precision mAP@[.5,.95] and (mAP@.5) for MS-COCO and VOC. The $\{ AP,AP_{50},AP_{75},AP^{Small},AP^{Middle},AP^{Large}$\} is reported for MS-COCO and $AP_{50}$ for VOC.

\textbf{Synthetic Noisy Dataset.}
Following \cite{oamil}, 
We simulate noisy bounding boxes by perturbing clean boxes from the original annotations. The details are in the appendix.
We simulate various box noise levels ranging from $10\%$ to $40\%$ for the VOC and $\{20\%, 40\%\}$ for the MS-COCO.

\textbf{Implementation Details.}
We implement our method on FasterRCNN \cite{FasterRCNN} with ResNet50-FPN \cite{DBLP:resnet, DBLP:FPN} backbone, based on MMDetection \cite{mmdetection}. All settings of our method and previous methods employ FPN for fair comparison. Similar to the default setting of object detection on MS-COCO, the stochastic gradient descent \cite{DBLP:SGD} algorithm is used to optimize on 1x training schedule. The batch size is two images per GPU on 8 GPUS. For the VOC dataset, the batch size is two images per GPU on 2 GPUS. The performance we report is on a single scale (1333 * 800 for MS-COCO and 1000 * 600 for VOC).

\begin{table}[tb!]
    \centering
    % \scriptsize
    \resizebox{0.47\textwidth}{!}{
    \begin{tabular}{l|c|c|c|ccc}
    \specialrule{0.13em}{0pt}{0pt}
    Methods ({$w/o$ SISD}) & AP & AP$_{50}$ & AP$_{75}$ &AP$^s$ &AP$^m$  &AP$^l$\\
    % \hline
    % \multicolumn{6}{l}{$w/o$ SISD:} \\
    \hline
     Neighborhood Sampler      & 24.6 & 52.0 &20.1& 14.3& 28.2 & 31.9  \\
     SPSD (II) $w/o$ weighted   &  26.0 & 53.3 &22.5& 15.6 & 29.4 &33.4 \\
     SPSD (II)  $w/$ weighted    &26.3 & 53.4 &22.5 & 15.6 & 29.3 & 33.8\\
    %  SPSD (I+II)$w/o$ weighted      &\\
     SPSD (I+II) $w/$ weighted      &\textbf{27.2} &\
     \textbf{53.7} &\textbf{24.7}& \textbf{15.9} & \textbf{30.3} & \textbf{35.2} \\
    %  \hline
    % \multicolumn{6}{l}{$w/o$ SISD:} \\
    % \hline
    % SPSD $w/$weighted loss  &   \\
    %  SPSD $w/o$ weighted loss     &\\
    %  SPSD only in stage II &   \\
    
    %  Hand-sampling     &\\
    %  Hand-sampling $w/o$ SISD  $w/$ re-train   & 26.52 & 53.97 & 15.66 & 30.34 & 33.77  \\
    %  SSD-Det $w/$ SEM    & &  &&&\\
    %  Re-sampling $w/$ SEM     &&&&  \\
    %  SSD-Det $w/$ SEM $w/$ re-train     & &  &&&\\
    %  Re-sampling $w/$ SEM $w/$ re-train      &&&&  \\
     \specialrule{0.13em}{0pt}{0pt}
    \end{tabular}}
        %\vspace{-5pt}
    \caption{Different setting of SPSD.}
    \label{tab:SPSD abla}
    %\vspace{-10pt}
\end{table}

\begin{table}[tb!]
    \centering
    \resizebox{0.47\textwidth}{!}{
    % \scriptsize
    \begin{tabular}{l|c|c|c|ccc}
    \specialrule{0.13em}{0pt}{0pt}
    ORE Strategies of SISD  & AP &  AP$_{50}$ & AP$_{75}$ &AP$^s$ &AP$^m$  &AP$^l$ \\
    \hline\
    $w/o$ SISD &27.2 & 53.7 & 24.7     &  15.9     & 30.3     & \textbf{35.2} \\
    SISD $w/o$ ORE &27.3 &53.3 &25.7&16.9&30.2&35.0 \\
    \hline
  
    + subtract  & 27.2 & 53.8 & 24.6 & 15.7 & 30.4 & 35.0 \\
    + concatenate  &  27.2 & 54.0 & 24.5 & 16.1 & 30.2& 34.8  \\
    % atten   \\
    + add &\textbf{27.6}  &\textbf{53.9} & \textbf{26.0} & \textbf{16.0} & \textbf{31.0} & 34.9  \\
    + add $w/$ shared fcs&23.0&49.9&17.6 & 12.6 & 25.3 & 30.4 \\
    % add $w/$ weighted & 27.52 & 53.59 & 25.83 &15。80 &30.58 &35.13 \\
    \specialrule{0.13em}{0pt}{0pt}
    \end{tabular}
    }
    %\vspace{-5pt}
    \caption{Different ORE strategies of SISD.}
    \label{tab:SISD}
    %\vspace{-10pt}
\end{table}

\begin{table}[t]
\begin{center}
    \resizebox{0.48\textwidth}{!}{
    \begin{tabular}{c|c|c|c|c}
    \specialrule{0.13em}{0pt}{0pt}
    % \multirow{2}[2]{*}{Num. of SPSD}
    Num.&0&1&2&3
    % &\multicolumn{2}{c|}{0}
    % &\multicolumn{2}{c|}{1}
    % &\multicolumn{2}{c|}{2 (ours)}
    % &\multicolumn{2}{c|}{3}
    % &\multicolumn{2}{c}{4}
    \\
    % &$AP$&$AP_{50}$&$AP$&$AP_{50}$&$AP$&$AP_{50}$&$AP$&$AP_{50}$ \\
    \hline
     AP/AP$_{50}$       & 24.6 / 52.0 &  26.3 / 53.4&\textbf{27.2 / 53.7}& 27.0 / 53.1\\
    % 1   & 26.29& 53.35& 22.51& 15.59& 29.25& 33.8 \\
    % 2 (our)   & 27.18& 53.74 &24.74 &15.86& 30.30& 35.16 \\
    % 3         & 27.03 & 53.13 & 25.33 &15.88&30.07&34.47\\
  \specialrule{0.13em}{0pt}{0pt}
    \end{tabular}
    }
\end{center}
%\vspace{-15pt}
\caption{Number of SPSD module.}
% %\vspace{-15pt}
\label{rebuttal:tab:spsd num}
%\vspace{-18pt}
\end{table}

\subsection{Comparison with State-of-the-Art}
We compare our method with several state-of-the-art approaches \cite{KLloss,DBLP:co-teaching,SD-LocNet,FreeAnchor,oamil} on MS-COCO and VOC 2007 datasets. We denote Clean-FasterRCNN and Noisy-FasterRCNN as FasterRCNN models trained under clean (original annotations) and noisy annotations with the default setting, respectively.

\textbf{MS-COCO Dataset.} Table \ref{tab:coco-table} shows the comparison results on the MS-COCO.
% noise decrease
Inaccurate bounding box annotations significantly deteriorate the vanilla Faster R-CNN's detection performance. 
Co-teaching and SD-LocNet only slightly improve the detection performance, especially under 40\% box noise. That indicates that small-loss sample selection and sample weight assignment can not tackle noisy box annotations well. KL Loss slightly improves the performance under 20\% and 40\% box noise.
By treating an object as a bag of instances, OA-MIL is somehow robust to noisy bounding boxes and performs better than other methods. Nevertheless, the previously-mentioned label assignment bag construction limits its ability to handle heavy noise. 
Our approach is more robust to noisy bounding boxes. It outperforms other methods by a large margin under high box noise levels and significantly boosts the detection performance across all metrics.
    For example, under 40\% box noise, the end-to-end SSD-Det achieves 27.6 $AP$ and 53.9 $AP_{50}$, 9.0 and attains 11.3 point improvement compared with state-of-the-art method OA-MIL, respectively. Also, through re-training on FasterRCNN, the performance further reaches 29.3 $AP$ and 54.8 $AP_{50}$. With the backbone of ResNet-101, the performance achieves consistent improvement. On MS-COCO test set, our method also achieves state-of-the-art performance.
\begin{figure*}[ht]
  \centering
  \includegraphics[width=\linewidth]{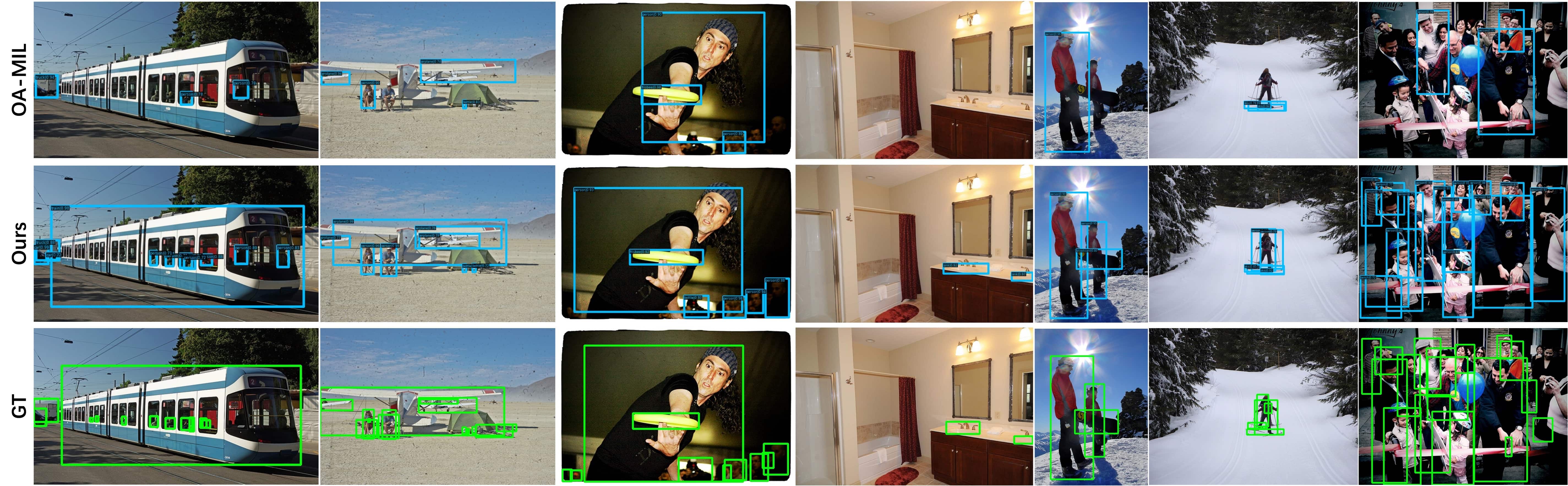}
     %\vspace{-15pt}
   \caption{Qualitative detection results on COCO validation. Previous methods miss objects and face part prediction problems. Our method misses fewer objects, and the bounding box quality is better, especially for small or overlapped objects.}
   \label{fig:vis_res}
   %\vspace{-13pt}
\end{figure*}
% :0.2 0.4

\textbf{VOC 2007 Dataset.} Table \ref{tab:voc-table} shows the comparison results on the VOC 2007 test set.
Co-teaching, SD-LocNet and KL Loss, can not address inaccurate bounding box annotations well. OA-MIL improves the performance on different noisy datasets.
Our approach obtains further improvements
to 77.10, 74.80, 71.50, 66.90 $AP_{50}$ on 10\%, 20\%, 30 \% and 40 \% noisy box datasets, respectively.

\subsection{Ablation Study and Analysis}
To further analyze SSD-Det’s effectiveness and robustness,
we conduct more experiments on COCO val set 
if there are no other instructions. 
Except for Table~\ref{tab:ablation on modules}, the noise level of these experiments is 40$\%$.

\textbf{Ablation of Modules.} Ablation study of each component in our approach is given in Table \ref{tab:ablation on modules}, including:
(i) Different stages of our basic box refiner. \ie training object detector without the stage II (2-Ref), where the pseudo boxes predicted by the stage I are served as the supervision for training a parallel detector.
(ii) SPSD, \ie training without SPSD, where the object-bag is constructed directly by neighborhood sampling around the noisy ground-truth or the predicted pseudo boxes of the stage I.
(iii) SISD.
(iv) Re-Train with FasterRCNN (Re-Train).

\begin{figure}[t]
  \centering
  \includegraphics[width=1.0\linewidth]{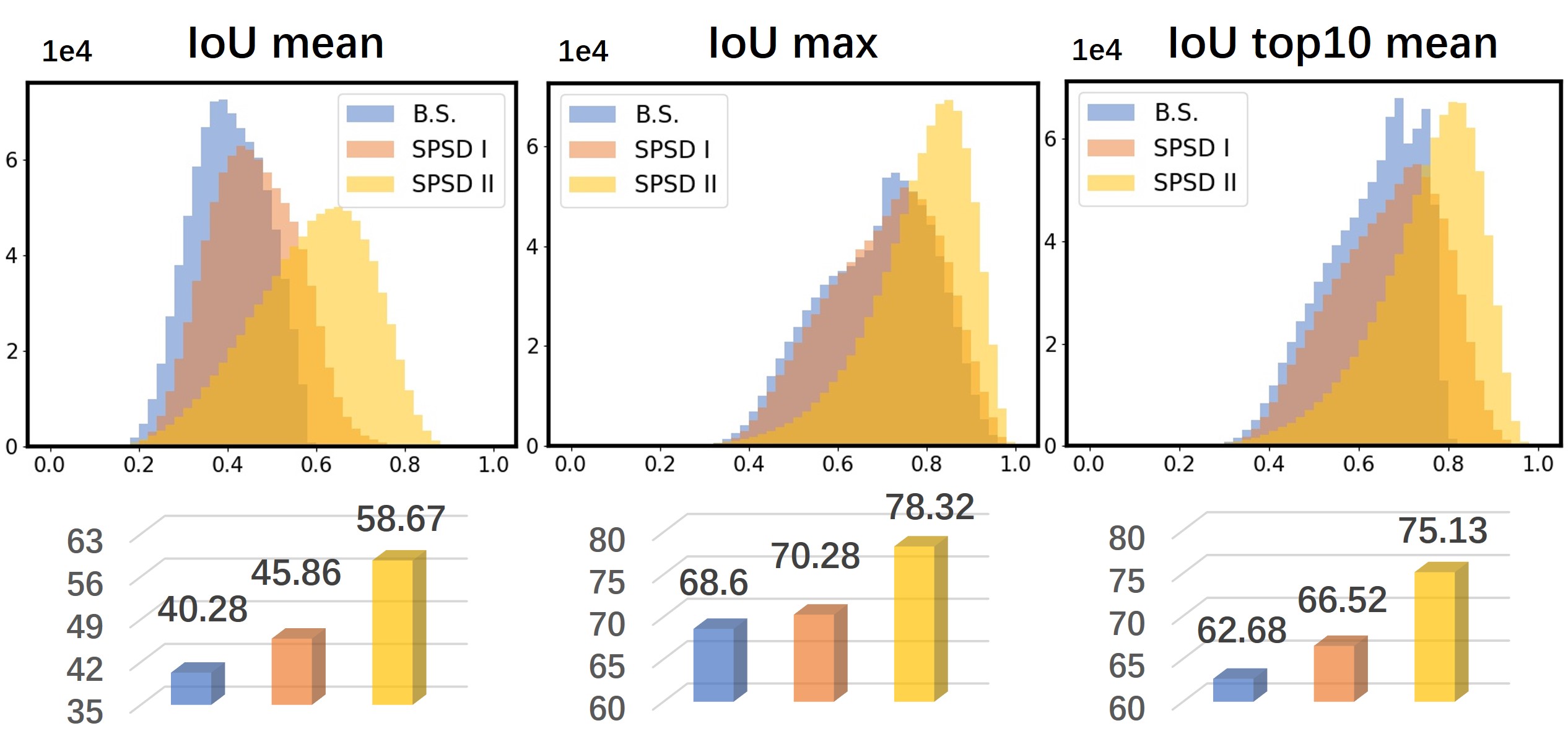}
     %\vspace{-15pt}
   \caption{Bag quality (IoU of proposals with GT) of construction in SSD-Det. B.S. (blue) means neighborhood sampler. SPSD I (orange) denotes single SPSD adopted. SPSD II (yellow) is two SPSD and interactive structure adopted. SPSD II significantly improves the quality.}
   \label{fig:vis_bag}
   %\vspace{-15pt}
\end{figure}

\textbf{Effectiveness of SPSD.}
SPSD further improves the detection performance on the MS-COCO, especially under high box noise levels, \eg under 40\% box noise level, SPSD boosts the performance
% from 56.0 to 63.3 on the VOC dataset and 
from 24.6 to 27.2, as shown in Table~\ref{tab:ablation on modules} (row 3).
In Table~\ref{tab:SPSD abla}, we conduct further ablation on SPSD. With SPSD bag construction only in stage II, the performance increases by 1.4 $AP$. The performance further improves with the proposal score of stage I as weights. With SPSD in all stages, the $AP$ reaches 27.2. Fig.~\ref{fig:vis_bag} shows the bag quality. With SPSD, the mean IoU increases from 40.3 to 58.7 and the max and top-10 IoU increase to 78.3 and 75.1, which indicates a better upper bound of proposal selection. More high-quality proposals bring better optimization and easier proposals selection.

\begin{table}[tb!]
    \centering
    \resizebox{0.47\textwidth}{!}{
    \scriptsize
    \begin{tabular}{l|c|c|c}
    \specialrule{0.13em}{0pt}{0pt}
    Methods & 
    Box Refiner+Re-Train  &SSD-Det&SSD-Det+Re-Train \\
    \hline
    AP/AP$_{50}$ & 29.0 / 54.4&27.6/53.9 & \textbf{29.3 / 54.8}      \\
    \specialrule{0.13em}{0pt}{0pt}
    \end{tabular}}
        %\vspace{-5pt}
    \caption{Comparisons of end-to-end and Re-Train.}
    \label{tab:Re train}
    %\vspace{-10pt}
\end{table}

\begin{table}[t]
\begin{center}
\setlength\tabcolsep{5pt}
\scriptsize
    \resizebox{0.48\textwidth}{!}{
    \begin{tabular}{l|cc|cc|ccc}
    \specialrule{0.13em}{0pt}{0pt}
    \multirow{3}[2]{*}{Detectors}&\multicolumn{2}{c|}{\makebox[0.07\textwidth][c]{Clean-supervised}}  &\multicolumn{4}{c}{Noise-supervised}           \\
    \cline{2-7}
    &\multicolumn{2}{c|}{} &\multicolumn{2}{c|}{}  &\multicolumn{2}{c}{($w/$ ours)} \\
    % \cline{2-8}
    &$AP$&\multicolumn{1}{c|}{$AP_{50}$}&$AP$&\multicolumn{1}{c|}{$AP_{50}$}&$AP$&\multicolumn{1}{c}{$AP_{50}$} \\
    \hline
    % \hline
    % \multicolumn{7}{l}{40\% Box Noise Level}\\
    % \hline
    FasterRCNN        & 37.9 &58.1& \textbf{10.3} &\textbf{28.9}&29.3 &54.8\\
    % SparseRCNN   & 27.20 & 50.50& 26.90&21.50&34.30&26.90\\
    SparseRCNN~\cite{DBLP:sparsercnn}$^{\dagger}$   &45.0&64.1& 6.0&20.3& 34.3&60.2\\
    \makebox[0.10\textwidth][l]{De-DETR~\cite{DBLP:DeformableDETR}$^{\dagger}$} &\textbf{46.8}&\textbf{66.3}&5.0&16.9&\textbf{35.2} & \textbf{60.9}\\
    % xxx         & 37.42 & 47.26 & 47.72 \\
    \specialrule{0.13em}{0pt}{0pt}
    % \hline
    % \multicolumn{7}{l}{20\% Box Noise Level}\\
    % \hline
    % FasterRCNN        & 37.90&58.10&30.40&31.40&34.37&57.34\\
    % % SparseRCNN   & 27.20 & 50.50& 26.90&21.50&34.30&26.90\\
    % SparseRCNN   & \\
    % Deformable DETR   &\\
    % \hline
    \end{tabular}
    }
\end{center}
%\vspace{-15pt}
\caption{Experiments on advanced detectors. De-DETR is Deformable DETR. ${\dagger}$ uses multi-scale data augment. '$w/$ ours' means using our method under noisy supervision.}
% %\vspace{-15pt}
\label{rebuttal:tab:recent detectors}
%\vspace{-10pt}
\end{table}

\textbf{Number of SPSD.} As shown in
Table~\ref{rebuttal:tab:spsd num}.
When adding 3 SPSD, performance drops slightly, probably due to the accumulation of errors outweighing the performance gain from extra stages. Hence, 2 SPSD is our default setting.

\textbf{Effectiveness of SISD.}
SISD is designed to select object-aware proposals in box selection. Under 40\% and 20\% box noise, the detection performance improves from 27.2 to 27.6 and 33.0 to 33.6, which verifies the effectiveness of the module, shown in Table~\ref{tab:ablation on modules}. We also study the strategies of ORE in SISD (Table~\ref{tab:SISD}). The minus or concat on object feature $\mathbf{F}_j^+$ and proposal feature $\mathbf{F}_j$ do not work. 
With add strategy, the performance is 27.60. 
If SISD shares the two fc layers, the performance drops to 22.99 since the optimization goals are contradictory (Identity distinguishes objects in the same category). If we directly use the RoI feature without ORE, the performance drops to 27.32 $AP$, verifying the effectiveness of the object relevance strategy.

\textbf{Affect of Re-Train.}
As most WSOD methods do, we re-run the experiments by training a fully supervised detector for better performance.
We find that if the SSD-Det only trains the refiner and uses the pseudo label to train the FasterRCNN, the result is good but lower than re-train after the end-to-end training given in Table~\ref{tab:Re train} (row 1). This is because joint training is beneficial for box refinement.

\textbf{Experiments on Advanced Detectors.} We re-train recent detectors, \eg SparseRCNN and Deformable DETR, under the boxes refined by our method. Table~\ref{rebuttal:tab:recent detectors} verifies that our method achieves consistency improvement.

\textbf{The computational cost discussion.} Similar to OA-MIL, our method adds SSD head to Faster R-CNN for auxiliary training to refine the noisy annotation and the head is not used during inference, the calculation cost during inference is same to standard Faster R-CNN.

\begin{table}[!t]
% \begin{table}[t]
\setlength\tabcolsep{3pt}
\begin{center}
    \resizebox{0.48\textwidth}{!}{
    \begin{tabular}{l|cccc|cccc|cccc}
    \hline
    \multirow{2}{*}{Methods}&\multicolumn{4}{c|}{Drift Rate (\%)}&\multicolumn{4}{c|}{Group Rate (\%)}&\multicolumn{4}{c}{Part Rate (\%)}\\
    % \hline
    &all&s&m&l&all&s&m&l&all&s&m&l\\
    \hline
    OA-MIL\cite{oamil}&15.1&17.8&17.1 &7.4
    &6.7&2.8&3.4&1.4
    &2.8&3.4&2.7&2.3
    \\
    Ours&\textbf{1.5}&1.0&1.3&1.4
    &\textbf{1.7}&1.2&0.5&0.7
    &\textbf{1.0}&0.5&1.1&1.3
    \\
    \hline
    \end{tabular}
    }
\end{center}
%\vspace{-15pt}
\caption{Breakdown of different problems during refinement (COCO under 40\% noise level). s, m and l mean small, middle and large scale.
}
%\vspace{-15pt}
\label{rebuttal:tab:drift}
% %\vspace{-5pt}
% \end{table}
\end{table}

\begin{table}[t]
    %\vspace{-5pt}
    \centering
    \resizebox{1\linewidth}{!}{
    \begin{tabular}{lc|cccc|cccc}
    \specialrule{0.13em}{0pt}{0pt}
    \multirow{2}{*}{Dataset}  &\multirow{2}{*}{our}& \multicolumn{4}{c|}{Quality (Average IoU)} & \multicolumn{4}{c}{Frequency (\%)} \\%&  \multicolumn{2}{c}{Good} 
      && All & Part  &  Oversize & Shift    & Reliable  & Part  &  Oversize & Shift \\
    \specialrule{0.13em}{0pt}{0pt}
    \multicolumn{9}{c}{Detector as Annotator} \\
    \hline
    \multirow{2}{*}{Objects-F}&&44.3 &24.1 & 33.6& 13.3
    &40.1&28.4 & 9.8 &21.7\\
    &\checkmark&47.0& 30.8&41.8 & 16.8
    &49.2& 23.1& 8.0 & 19.4 \\
    \hline
    % \hline
    \multirow{2}{*}{COCO-F}&& 45.1&25.4&33.3&15.7
    &40.0 &27.5&12.0&20.5\\ %Recall 84.1
    &\checkmark&48.2&32.1&40.6&19.6
    &49.5 &22.2&10.0& 18.3\\
    \hline
   \multicolumn{9}{c}{Point-based Annotator}\\
    \hline
    \multirow{2}{*}{COCO-P}&& 55.6&30.4&25.0&29.8
    &65.6&9.5&23.2&1.7 \\
    &\checkmark&65.2&46.7&36.4&40.9
    &74.9&5.9&18.5&0.6 \\
    \specialrule{0.13em}{0pt}{0pt}
    \end{tabular}
    }
    %\vspace{-10pt}
    \caption{Analysis of noisy annotations types and quality. }
    %\vspace{-15pt}
    \label{tab:noisy type}
\end{table}

%\vspace{-2pt}
\subsection{Visualization and Discussion.}
Fig.~\ref{fig:vis_res} shows that OA-MIL faces missing instances and grouping instances issues for small or overlapped objects (as mentioned in \cite{oamil}), while our method still works well.
For a better intuitive understanding of SISD and SPSD, we visualize the bag construction quality in Fig.~\ref{fig:vis_bag}
Then, we makes noise types breakdown of 'Drift', 'Group' and 'part dominance' issues. We give the definition of $IoU$, $IoG$ and $IoD$:
%\vspace{-5pt}
\begin{equation}\scriptsize
    \begin{aligned}
      IoU = \frac{A(I)}{A(D)+A(G)-A(I)}, IoG = \frac{A(I)}{A(G)}, IoD = \frac{A(I)}{A(D)} \\
    \end{aligned}
%\vspace{-6pt}
\end{equation}
\noindent where A(*) is area of box *, D and G are refined box and gt box respectively, and I is insertion between D and G. We statistically count the proportion of three noise types of 'bad' refined boxes (having small IoU with gt) in Table \ref{rebuttal:tab:drift}:
% We divide the 'bad' refined boxes () into 3 types for statistics:
(i) Drift: 'bad' refined box has a higher IoU with another nearby object.
(ii) Group: 'bad' refined box has high IoG with multiple objects.
(iii) Part: 'bad' refined box has a high IoD.
Table \ref{rebuttal:tab:drift} shows quantitative results for each noise type of baseline and ours. The drift, group, part problems reduce from 15.1\%, 6.7\%, 2.8\% to 1.5\%, 1.7\%, 1.0\% , respectively, demonstrating our
% effectiveness and 
improvement.

\textbf{Experiments on Real-life Noisy Annotations.}
Real-life noisy annotations stem from: low-quality data (e.g., occlusion, blur), human annotator errors and automatic machine annotator limitations. 
Noise from human errors is quite subjective, since differences between annotators.
For a more objective analysis, noisy annotations from machine annotator is used for experiments.
Without loss of generality,
Faster R-CNN well-trained on MS-COCO was applied to Objects365 images, yielding Objects-F dataset,% (Recall 80.6) with 10,000 images and 137,818 noisy annotations.
and to COCO-val images, producing COCO-F dataset. 
P2BNet~\cite{p2b}, a point-based annotator, was used on COCO-val images with point annotations, generating COCO-P dataset.
SSD-Det effectively improves low-quality boxes.
As shown in Table~\ref{tab:noisy type},% shows the analysis of annotations quality and frequency. 
with SSD-Det's refinement, the average IoU increases for Objects-F (from 44.3 to 47.0), COCO-F (from 45.1 to 48.2) and COCO-P (from 55.6 to 65.2) datasets. Further, the proportion of reliable annotations increases, and noise categories' frequency (Part, Oversize, and Shift) decreases for all datasets.

%\vspace{-8pt}
\section{Conclusion}
%\vspace{-7pt}

This paper investigates problems during refinement caused by solely using category information to select proposals. We also propose SSD-Det to mine spatial information in a self-distillation fashion. SSD-Det introduces the SPSD module to learn semantic-spatial correspondence knowledge with neighborhood sampler and an interactive structure to combine spatial information and category information, thus producing a high-quality proposal bag.
SISD in SSD-Det is utilized to improve the proposal selection procedure by integrating object-relevant spatial confidence. Complete ablations on multiple datasets verify the effectiveness of SSD-Det.

{\small
\bibliographystyle{ieee_fullname}
\bibliography{paper_supple}
}
\clearpage

\begin{appendices}
\begin{center}{\bf \Large Appendix}\end{center}%\vspace{-2mm}

\begin{figure*}[t]
  \centering
  \includegraphics[width=.8\linewidth]{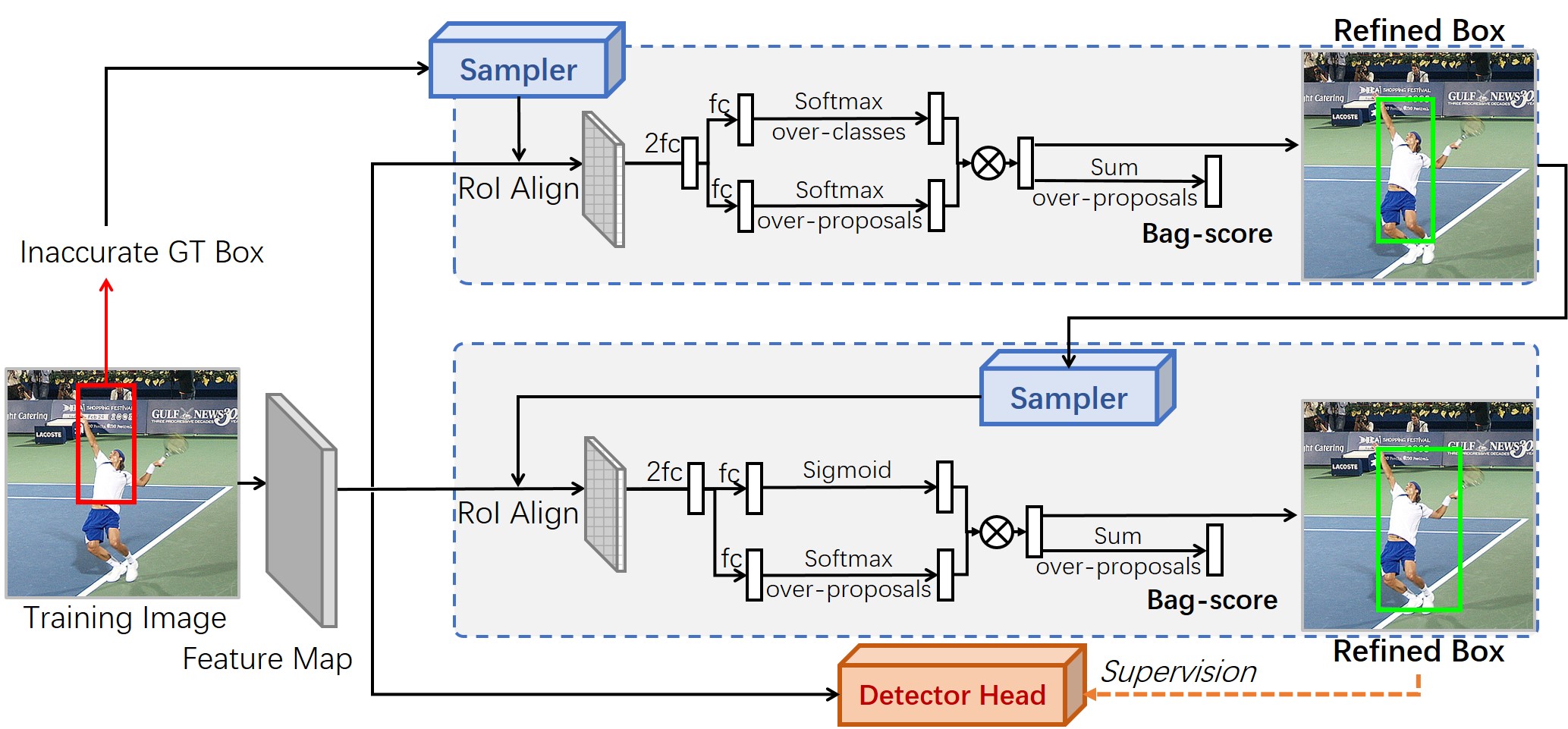}
   \caption{The basic box refiner.}
   \label{fig:app_basic}
\end{figure*}

\begin{figure*}[t]
  \centering
  \includegraphics[width=\linewidth]{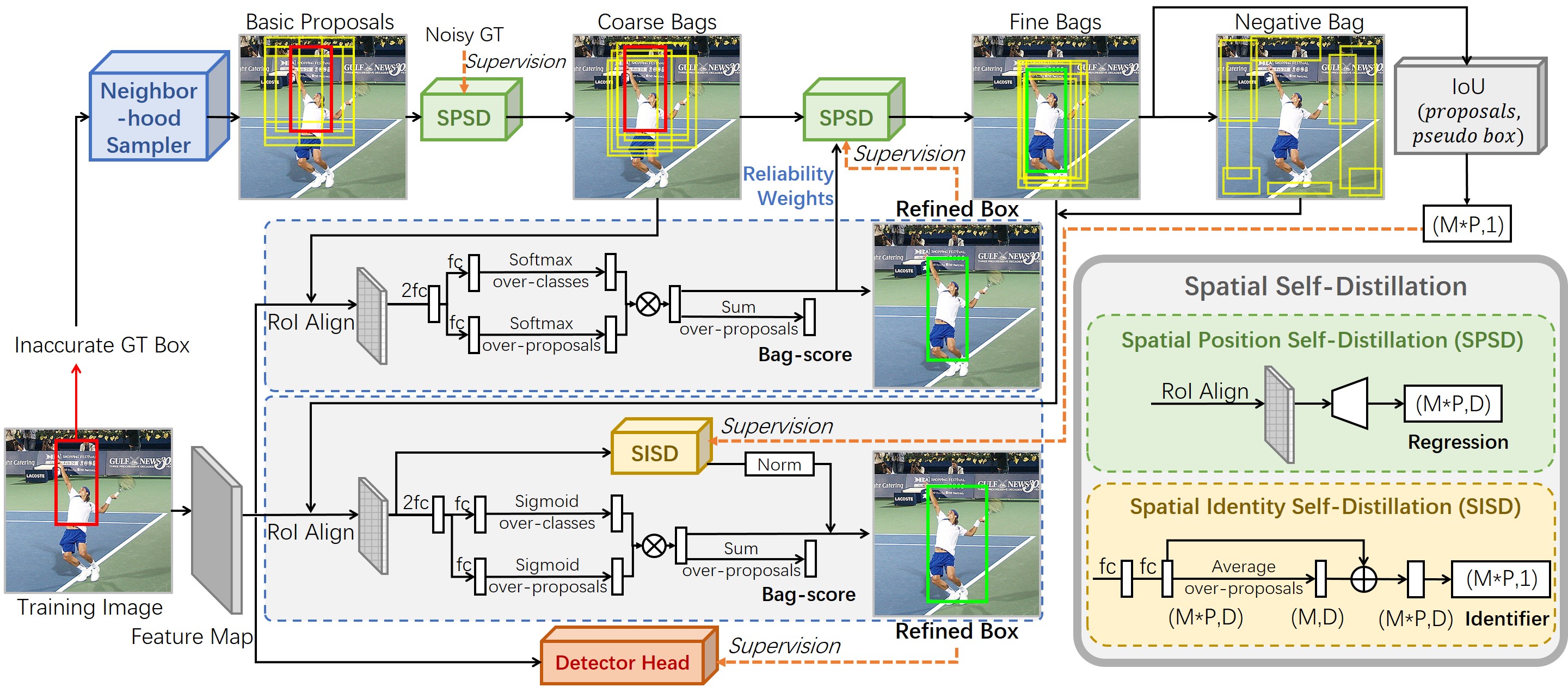}
   \caption{SSD-Det (SPSD shares backbone with the detector).}
   \label{fig:app_ssddet}
\end{figure*}

\begin{figure*}[t]
  \centering
%   \fbox{\rule{0pt}{0.5in} \rule{0.9\linewidth}{0pt}}
  \includegraphics[width=\linewidth]{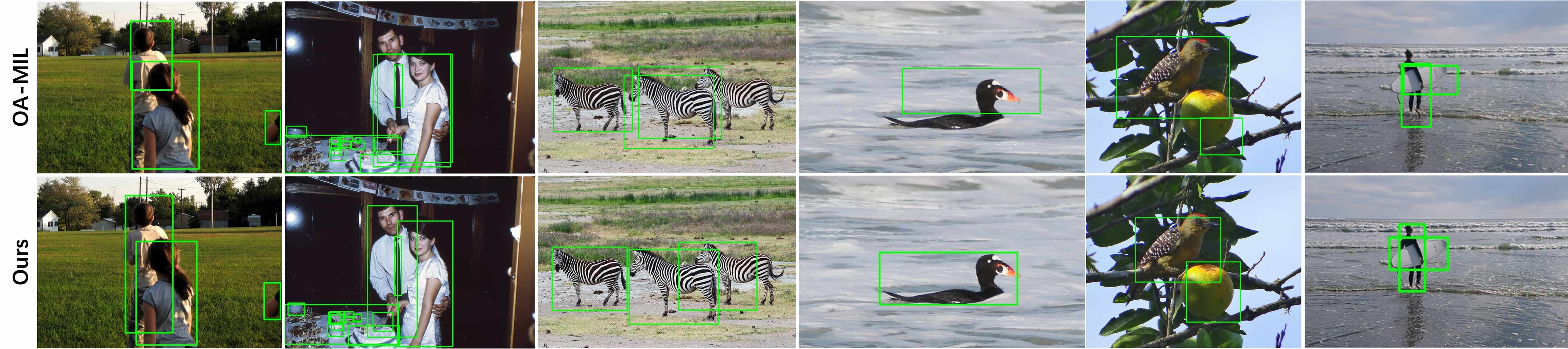}
     %\vspace{-20pt}
   \caption{Examples of the refined instances (MS-COCO train set under 40\% noise level).}
   \label{fig:app_refine}
  \vspace{-13pt}
\end{figure*}

\section{Codes}
The code of this paper is also included as a zip file (ssd-det.zip) in the supplementary. The submitted version contains training codes on MS-COCO\cite{coco} and VOC\cite{DBLP:VOC}. The details are given in README.md in the zip file.

\section{Details of SSD-Det Deployment}
% \textbf{Basic Box Refiner.}

\textbf{ Structure Details.}
Fig.~\ref{fig:app_basic} depicts the detailed structure of the basic box refiner,
while Fig.~\ref{fig:app_refine} depicts the detailed structure of our SSD-Det.

\textbf{Implementation Details.}
ResNet-50 is used as the backbone network unless otherwise specified, and FPN is adopted for feature fusion.
The mini-batch is 16 images; all models are trained with 8/2 GPUs and 2 images per GPU for MS-COCO/VOC. The training epoch numbers are set as 12, and the learning rate is set as 0.02/0.002 and decays by 0.1 at the 8-th and 11-th epoch for MS-COCO/VOC.  In default settings, the backbone is initialized with the pre-trained weights on ImageNet and other newly added layers are initialized with Xavier. 
% The sampling radius $R$ is set as 8/7/5 for COCO/DOTA/SeaPerson by defalut.
In 40\% noise rate in MS-COCO, the original settings of basic sampling are:$(v\cdot s) \in \{0.7, 0.8, 1, 1.2, 1.3\}$, $(v/s) \in \{0.7, 0.8, 1, 1.2, 1.3\}$ and $(o_x,o_y) \in \{(0,0), (2,0), (0,2), (-2,0), (-2,-2)\}$ is used to jitter the centre position. Those are set the half for the 20\% noise rate dataset. The settings in VOC are the same and adaptively changed for other noise rate datasets. In negative sampling, we randomly sample 500 boxes, filter out those which have high IoU (0.3) with all positive proposals and obtain the final negative sample set $\mathcal{N}$. The loss weights are set as  $\alpha_1, \alpha_2, \alpha_3$ and $\alpha_4$ are set as 1, 0.25, 0.25 and 4, respectively, without much hyper-parameter tuning.

\textbf{Synthetic Noisy Dataset.}
Following \cite{oamil}, 
we simulate noisy bounding boxes by perturbing clean boxes from the original annotations. Specifically, $cx$, $cy$, $w$, and $h$ denote an object's the center $x$ coordinate, center $y$ coordinate, width, and height, respectively. We simulate an inaccurate bounding box by randomly shifting and scaling the box as follows:
\begin{equation}\small
\begin{cases}\hat{c} x=c x+\Delta_x \cdot w, & \hat{c y}=c y+\Delta_y \cdot h \\ \hat{w}=\left(1+\Delta_w\right) \cdot w, & \hat{h}=\left(1+\Delta_h\right) \cdot h\end{cases}
\label{Eq: noisy box generation}
\end{equation}
where $\Delta x$, $\Delta y$, $\Delta w$, and $\Delta h$ obey the uniform distribution $U(-r,r)$, and $r$ is the box noise level. For example, when $r$ = $40\%$, $\Delta x$, $\Delta y$, $\Delta w$, and $\Delta h$ are in the range of $(-0.4,0.4)$.
We simulate various box noise levels ranging from $10\%$ to $40\%$ for the VOC dataset and $\{20\%, 40\%\}$ for the MS-COCO dataset. Eq. \ref{Eq: noisy box generation} is conducted on every bounding box in the training dataset.  

\section{Details of Average IoU}

\textbf{Average IoU} is the evaluation metric of the performance of dataset refine, and the higher average IoU means the better performance. Table \ref{tab append: Mean IoU of pseudo box} shows that the quality of dataset refinement is greatly improved after OA-MIL solves the drift problem. By simply filtering out the pseudo box with $IoU=0$, the performance of OA-MIL improves from 47.6 to 54.4. Further, once filtering out the pseudo box with $IoU=0$, the performance of OA-MIL improves from 47.6 to 54.4. 
If the pseudo frame with $IoU\leq 0.5$ is filtered out, OA-MIL's refinement performance is close to ours.
If only the proposals whose IoU with GT is greater than 1e-5 are counted (second line), the average IoU of OA-MIL is greatly increased, meaning lots of extremely low-quality refined results, while IoU of our SSD-Det remains essentially unchanged.

\begin{table}[ht]
    \centering
    \resizebox{0.47\textwidth}{!}{
    \begin{tabular}{l|cccc}
    \specialrule{0.13em}{0pt}{0pt}
       \multirow{2}{*}{Methods} & \multicolumn{4}{c}{Average IoU} \\
      &IoU$\ge$0 &IoU \textgreater0 & IoU \textgreater0.3&IoU \textgreater0.5 \\
       \hline
       (40\% Noise Level)                 &46.4&-&-&- \\ 
       OA-MIL\cite{oamil}  &47.6&54.4& 57.1&67.5  \\ 
    %   Basic refiner&62.2&64.2&69.1&62.2\\
       SSD-Det  &65.1&65.1& 67.7 &72.7 \\ 
        % &- &&69.1 (+6.9)& \\ 
       \specialrule{0.13em}{0pt}{0pt} 
    \end{tabular}}
        %\vspace{-8pt}
    \caption{The average IoU of different methods' refined boxes with clean GT on MS-COCO under 40\% Noise Level.}
    \label{tab append: Mean IoU of pseudo box}
    %\vspace{-10pt}
\end{table}

\section{Qualitative Results}

\begin{figure*}[ht]
  \centering
%   \fbox{\rule{0pt}{0.5in} \rule{0.9\linewidth}{0pt}}
  \includegraphics[width=.56\linewidth]{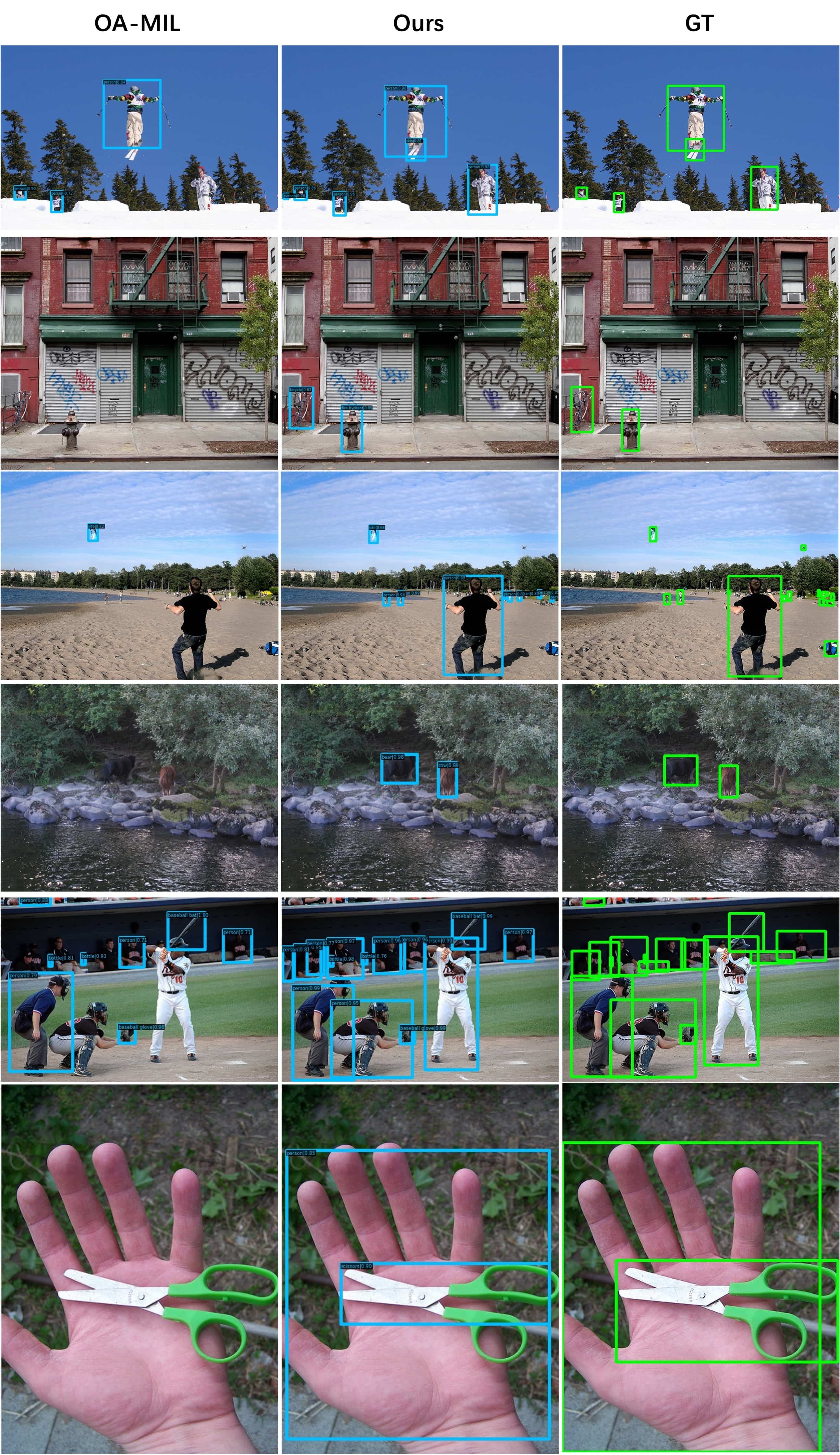}
    %  %\vspace{-20pt}
   \caption{Qualitative results on MS-COCO validation set.}
   \label{fig:app_res}
\end{figure*}

\textbf{Affect of Re-Train.}
As most WSOD methods do, we rerun the experiments by training a fully supervised detector, \eg Faster R-CNN or RetinaNet, to regress the object locations more precisely. As shown in Table \ref{tab:Re train}, we get a better result of 20.29 AP and 34.37 AP on 40\% and 20\% noise datasets. We also find that if the SSD-Det only trains the refiner and uses the pseudo label to train the FasterRCNN, the result is good but lower than re-train after the end-to-end training given in Table~\ref{tab:Re train} (row 1). This is because joint training is beneficial for box refinement.
% \subsection{Point Supervised Object Detection}

\begin{table}[hb!]
    \centering
    \resizebox{0.49\textwidth}{!}{
    \scriptsize
    \begin{tabular}{l|c|c|c|ccc}
    \specialrule{0.13em}{0pt}{0pt}
    Methods & AP &  AP$_{50}$ & AP$_{75}$ &AP$^s$ &AP$^m$  &AP$^l$ \\
    \hline
    Box Refiner+Re-Train     & 29.0 & 54.4 & 28.2 & \textbf{17.7} & 32.3 & 36.4 \\
    SSD-Det&27.6  &53.9 & 26.0 & 16.0 & 31.0 & 34.9 \\
    SSD-Det+Re-Train      &    \textbf{29.3}    & \textbf{54.8}      &  \textbf{29.0}     &    17.1   & \textbf{32.9}      &  \textbf{36.9}\\
    \specialrule{0.13em}{0pt}{0pt}
    \end{tabular}}
        % %\vspace{-5pt}
    \caption{Comparisons of end-to-end and re-train (40\% noise).}
    \label{tab:Re train abla}
    %\vspace{-10pt}
\end{table}

\textbf{Experiments on Different Detectors.} Experiments are conducted on ResNet50. We re-train the different detectors with corrected labels. Table~\ref{tab:different detectors} shows the detection results, verifying the robustness of our method.

\begin{table}[hb!]
    \centering
    \resizebox{0.47\textwidth}{!}{
    \scriptsize
    \begin{tabular}{l|c|c|c|ccc}
    \specialrule{0.13em}{0pt}{0pt}
        Detectors & AP &  AP$_{50}$  & AP$_{75}$ &AP$^s$ &AP$^m$  &AP$^l$ \\
    \hline                                  
      Faster R-CNN   & 29.3    & 54.8      &  29.0     &    17.1   & 32.9      &  36.9             \\
      RetinaNet   &28.6& 52.8 & 28.8  & 17.1  & 32.3 & 36.4   \\
      RepPoints   &28.6&53.7&28.0&16.8&32.0  & 37.0 \\
      Free-Anchor  &29.4&54.1&29.6&17.0&32.4&37.6   \\
      Sparse R-CNN &34.3&60.2&36.4&22.4&37.5&43.7\\
      Deformable-DETR&35.0&60.7&37.4&23.6&38.1&44.4\\
    \specialrule{0.13em}{0pt}{0pt}
    \end{tabular}}
    % %\vspace{-5pt}
    \caption{Different detectors for re-train (40\% noise).}
    \label{tab:different detectors}
    %\vspace{-10pt}
\end{table}

\textbf{Visualization.}
Fig. \ref{fig:app_refine} shows the refined boxes predicted by OA-MIL and our SSD-Det on the MS-COCO datasets with 40\% box noise.
We can observe that OA-MIL suffers from object drift, group prediction, part domination problems.
Fig. \ref{fig:app_res} shows the qualitative results of the OA-MIL and our SSD-Det on the MS-COCO datasets with 40\% box noise.

\end{appendices}
\end{document}